\documentclass[10pt,twocolumn,letterpaper]{article}

\usepackage[pagenumbers]{cvpr} 

\usepackage{multirow}
\usepackage{booktabs}

\definecolor{cvprblue}{rgb}{0.21,0.49,0.74}
\usepackage[pagebackref,breaklinks,colorlinks,allcolors=cvprblue]{hyperref}
\usepackage[font=small,skip=5pt]{caption}

\def\paperID{17098} 
\def\confName{CVPR}
\def\confYear{2025}

\title{Implicit Neural Representation for Video and Image Super-Resolution}

\author{
Mary Aiyetigbo\footnotemark[1], \hspace{1em}
Wanqi Yuan\footnotemark[1],\hspace{1em}
Feng Luo, \hspace{1em}
Nianyi Li, \hspace{1em} \\
Clemson University\\
{\tt\small \{maiyeti,wanqiy,luofeng,nianyil\}@clemson.edu}
}

\begin{document}
\maketitle

\footnotetext[1]{Equal contribution.}

\begin{abstract}
We present a novel approach for super-resolution that utilizes implicit neural representation (INR) to effectively reconstruct and enhance low-resolution videos and images. By leveraging the capacity of neural networks to implicitly encode spatial and temporal features, our method facilitates high-resolution reconstruction using only low-resolution inputs and a 3D high-resolution grid. This results in an efficient solution for both image and video super-resolution.
Our proposed method, SR-INR, maintains consistent details across frames and images, achieving impressive temporal stability without relying on the computationally intensive optical flow or motion estimation typically used in other video super-resolution techniques. The simplicity of our approach contrasts with the complexity of many existing methods, making it both effective and efficient.
Experimental evaluations show that SR-INR delivers results on par with or superior to state-of-the-art super-resolution methods, while maintaining a more straightforward structure and reduced computational demands. These findings highlight the potential of implicit neural representations as a powerful tool for reconstructing high-quality, temporally consistent video and image signals from low-resolution data.
\end{abstract}   
\section{Introduction}
High-resolution (HR) images and videos are essential for numerous computer vision applications, including surveillance, medical imaging, and multimedia entertainment. However, capturing high-resolution data is often constrained by hardware limitations, bandwidth, and storage considerations. Consequently, super-resolution (SR) techniques, which aim to reconstruct HR content from low-resolution (LR) inputs, have become a critical area of research~\cite{park2003super}. 
Existing SR methods can be broadly categorized into single-image super-resolution (SISR) and video super-resolution (VSR). SISR methods focus on enhancing individual frames without leveraging temporal information~\cite{ledig2017photo, zhang2018rcan, liang2021swinir}. While these approaches have achieved impressive results, they often fail to produce temporally consistent outputs when applied to video sequences due to the lack of temporal modeling~\cite{caballero2017real}.

\begin{figure}[t]
    \centering
    \includegraphics[width=0.5\textwidth]{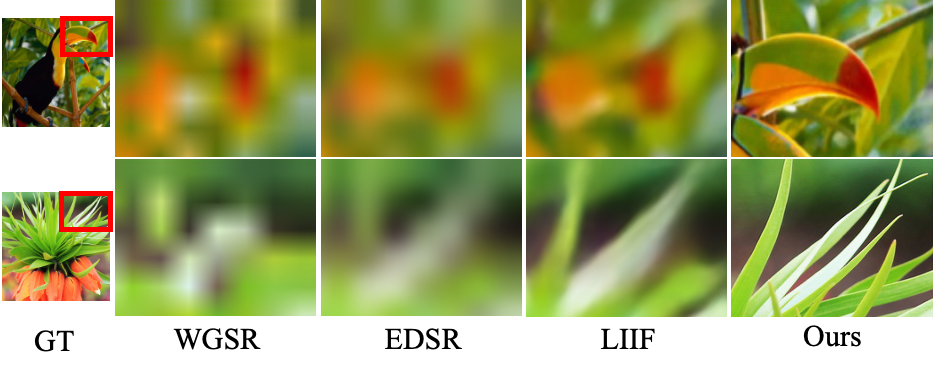}
    \caption{Visual comparison of image details in SOTA super-resolution. The leftmost column shows the ground truth, followed from left to right by the results of WGSR \cite{korkmaz2024training}, EDSR \cite{edsr}, LIIF \cite{chen2021liif}, and our method, with an upsampling scale of 32.}
    \label{detail}
\end{figure}

VSR methods exploit temporal correlations between consecutive frames to improve reconstruction quality and ensure temporal consistency~\cite{kappeler2016video, tao2017detail, wang2019edvr, chan2021basicvsr}. VSR approaches account for motion either by computing explicit motion estimation, using optical flow, to align frames before reconstruction~\cite{kappeler2016video, caballero2017real}. 
or implicitly motion through convolutions, dynamic filters, or transformer networks to capture long-range dependencies in both spatial and temporal dimensions, enhancing temporal consistency and reconstruction quality~\cite{liang2022vrt, liang2022recurrent, shi2022rethinking, xu2023video, tian2020tdan, wang2019edvr, jo2018deep, yang2025motion, lyu2024stadnet, zhou2024upscale, chen2024learning}.

\begin{figure*}[t!]
    \centering
    \includegraphics[width=0.9\textwidth]{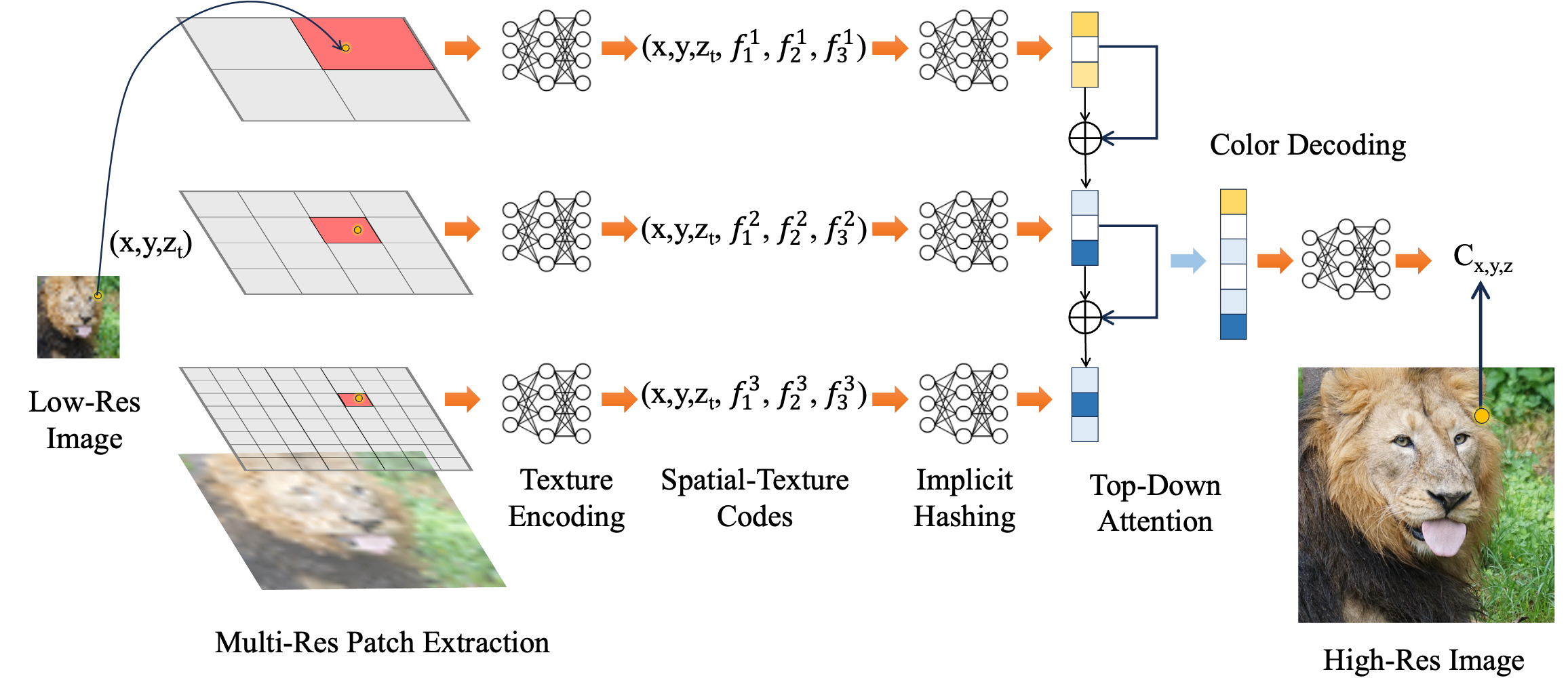}
    \caption{SR-INR pipeline for super-resolution. Local patches are extracted at multiple resolutions and processed by MLPs to generate feature vectors. These vectors are concatenated, refined via a top-down attention mechanism, and fed into an MLP to predict the RGB value, resulting in the super-resolved output.}
    \label{pipelien}
\end{figure*}

Implicit Neural Representations (INRs)\cite{mildenhall2020nerf, barron2021mipnerf, sitzmann2020siren, tancik2020fourier} have emerged as a powerful paradigm for continuous image and video representation. By representing signals as continuous functions parameterized by neural networks, INRs enable reconstruction at arbitrary resolutions without relying on discrete grids. This property makes them particularly suitable for super-resolution tasks, where flexible upsampling capabilities and the ability to model high-frequency details are essential.
Recent advancements have extended Implicit Neural Representations (INRs) to super-resolution tasks for images and videos \cite{chen2021liif, lee2024locality}. LIIF~\cite{chen2021liif} enables continuous upsampling using local implicit functions, while Bhat \etal~\cite{bhat2021deep} enhances modeling of complex structures with deep implicit functions.
NeRV~\cite{chen2021nerv} and HNeRV~\cite{han2022hnerv} model videos by mapping temporal indices to high-resolution frames, implicitly capturing temporal correlations. VideoINR~\cite{chen2022videoinr} represents videos as continuous space-time functions, allowing both spatial and temporal super-resolution at arbitrary resolutions and frame rates.
However, existing INR-based methods are typically designed specifically for either images~\cite{chen2021liif, bhat2021deep} or videos~\cite{chen2021nerv, chen2022videoinr}, limiting their flexibility when both modalities need to be addressed within a unified framework. In contrast, our approach is capable of handling both image and video super-resolution tasks. Inspired by Instant Neural Graphics Primitives (Instant-NGP)~\cite{mueller2022instant}, which efficiently represent scenes using multi-resolution hash encoding, we extend this concept to super-resolution tasks. 


In this paper, we propose a novel approach to super-resolution that leverages Implicit Neural Representations (INRs) and multi-resolution hash encoding for efficient high-resolution reconstruction from low-resolution videos and images. Our method, {SR-INR}, utilizes grid-based encoders with hierarchical representation learning to capture spatial and temporal features effectively. By implicitly encoding the input data and corresponding coordinates, our approach enables high-resolution reconstruction.



Our main contributions are as follows:

\begin{itemize} \item We propose hierarchical grid-based encoding that leverages multi-resolution hash encoding in ~\cite{mueller2022instant} to construct Implicit Neural Representations (INRs) for super-resolution tasks. \item Unlike prior methods that are tailored specifically for either images or videos, our framework is unified and capable of handling both image and video super-resolution. \item We provide extensive experiments and analyses to validate the effectiveness of our method. We demonstrate that our approach achieves competitive or superior results compared to state-of-the-art super-resolution methods. \end{itemize}
\section{Related Work}
\subsection{Image Super Resolution (ISR)}
The introduction of deep learning for ISR began with SRCNN by Dong \etal \cite{srcnn}, marking a significant milestone in super-resolution. Following this, numerous deep learning-based approaches were developed to enhance performance. Kim \etal \cite{vdsr} proposed the Very Deep Super-Resolution (VDSR) network, which utilized deeper architectures and residual learning to improve performance. Residual connections became essential in SR networks to address training difficulties in deep models. Ledig \etal \cite{srgan} employed generative adversarial networks (GANs) with a residual network (RESNET) \cite{resnet} backbone to produce more realistic textures. Lim \etal \cite{edsr} proposed the Enhanced Deep Residual Network (EDSR), improving efficiency by removing unnecessary layers such as batch normalization. Zhang \etal introduced the Residual Channel Attention Network (RCAN) \cite{rcan}, incorporating channel attention mechanisms to adaptively rescale features.
The success of vision transformers \cite{vit} and their ability to model long-range dependencies led to the use of attention mechanisms in ISR. Dai \etal proposed the Second-order Attention Network (SAN) \cite{san}, which captured feature interdependencies using second-order statistics. Liang \etal introduced SwinIR \cite{liang2021swinir}, utilizing Swin Transformers \cite{swint} with hierarchical feature maps and shifted windows to capture both local and global information effectively. Generative models, such as GANs and diffusion models, have also been highly effective for super-resolution tasks \cite{yu2024scaling, wu2024seesr}.

\subsection{Video Super Resolution (VSR)}
Video super-resolution (VSR) leverages temporal information from adjacent frames, presenting challenges like motion estimation and alignment. VSR methods can be broadly classified into sliding window and recurrent methods. Sliding window methods process consecutive frames together to reconstruct a high-resolution frame. \cite{kappeler2016video, caballero2017real, li2020mucan} relied on explicit motion estimation to align neighboring frames, which made them heavily dependent on accurate optical flow, prone to errors in complex scenes. To overcome this dependency, other approaches \cite{jo2018deep, tao2017detail, xue2019video, yi2019progressive} adopted implicit motion estimation. Deformable convolutions \cite{xu2021temporal, tian2020tdan, wang2019edvr} proved effective in capturing spatial and temporal dependencies without explicit motion estimation. The Video Restoration Transformer (VRT) \cite{liang2024vrt} further enhanced performance by capturing long-range dependencies.
Recurrent methods \cite{shi2015convolutional, chan2021basicvsr, chan2022basicvsr++, liang2022recurrent, haris2019recurrent, huang2017video, sajjadi2018frame, shi2022rethinking, yi2021omniscient} excel at capturing long-term temporal information by processing frames sequentially while maintaining a hidden state. These methods effectively handle complex motion and model temporal dependencies. 

\subsection{Implicit Neural Representation for Super-Resolution}
Implicit Neural Representations (INRs) have emerged as an effective approach for modeling continuous image and video data by representing signals as continuous functions parameterized by neural networks. This allows high-quality reconstruction and super-resolution at arbitrary scales. Chen \etal introduced the Local Implicit Image Function (LIIF) \cite{chen2021liif}, which learns continuous representations using local implicit functions, enabling upsampling to arbitrary resolutions. Bhat \etal proposed Deep Implicit Functions \cite{bhat2021deep} to handle complex image structures, while SIREN \cite{sitzmann2020siren} demonstrates that periodic activation functions help represent complex natural signals for super-resolution tasks. Tancik \etal presented Fourier Features \cite{tancik2020fourier} to enable networks capture fine details. 

Soh \etal introduced PIIFGAN \cite{soh2022piifgan}, leveraging periodic implicit functions for general image super-resolution across various scaling factors. An \etal proposed HyperINR \cite{an2021hyperinr}, using a hypernetwork to generate the weights of an INR conditioned on image content. For video super-resolution, Chen \etal proposed NeRV \cite{chen2021nerv}, which models videos by mapping frame indices to high-resolution frames without explicit pixel storage. Han \etal extended this with HNeRV \cite{han2022hnerv}, combining implicit and explicit representations. Chen \etal also introduced VideoINR \cite{chen2022videoinr}, modeling videos as continuous functions over space and time, allowing for both spatial and temporal super-resolution at arbitrary resolutions and frame rates. Lee \etal combined INRs with learned encoders for image super-resolution \cite{lee2022inr}, while Batzolis \etal proposed Implicit Diffusion Models \cite{batzolis2022implicit}, combining INRs with diffusion processes to generate high-resolution images from low-resolution inputs.

Our method draws inspiration from Instant-NGP \cite{mueller2022instant}, utilizing multi-resolution hash encoding for efficient neural representation. While Instant-NGP focuses on 3D scene reconstruction, we adapt its multi-resolution hash embedding approach to the super-resolution domain for both images and videos. This allows us to achieve efficient and high-quality super-resolution, generalizing across diverse 2D image and video data. Other works, such as DirectVoxGO \cite{sun2022direct}, TensoRF \cite{chen2022tensorf}, and Plenoxels \cite{yu2022plenoxels}, utilize similar encoding techniques for 3D scene representation and rendering. However, these methods are designed for 3D tasks and do not address image and video super-resolution. Our work differs by applying these encoding strategies to super-resolution tasks, demonstrating the versatility of grid-based and hash embedding techniques beyond 3D applications.

\section{Method}

In this section, we describe our proposed method, SR-INR, for image and video super-resolution. We start by defining the problem, then provide detailed explanations of each stage in our pipeline: texture encoding, implicit hashing, and top-down attention.

\subsection{Problem Definition}
The input to SR-INR is a batch of low-resolution frames $I_{lr} \in \mathbb{R}^{H_{lr} \times W_{lr} \times C}$, and a high-resolution grid $\mathcal{G}_{hr}$ specifying the locations of the high-resolution pixel coordinates. The output is the RGB color value corresponding to each high-resolution coordinate. Mathematically, we denote this as:
\begin{equation}
    I_{hr} = f(I_{lr}, \mathcal{G}_{hr}),
\end{equation}
where $I_{hr} \in \mathbb{R}^{H_{hr} \times W_{hr} \times C}$ represents the reconstructed high-resolution image.

\subsection{Texture Encoding}
The texture encoding process starts by extracting local patches from the low-resolution images based on a batch of resolution grids. Inspired by Instant-NGP \cite{mueller2022instant}, we adopt a strategy of using a multi-resolution grid to generate local feature representations. Specifically, given a high-resolution pixel coordinate, we extract multi-scale local patches from the low-resolution frames based on the resolution grid:
\begin{equation}
    P_i^r = g(I_{lr}, \mathcal{G}_{r}),
\end{equation}
where $P_i^r$ represents a local patch centered around coordinate $i$ at resolution level $r$. The size of the local patch is determined by the resolution grid $\mathcal{G}_{r}$.

We then predict a fixed-length feature code for each of these multi-resolution patches:
\begin{equation}
    \hat{\mathcal{F}}_i^r = f^r_\theta(P_i^r),
\end{equation}
where $\mathcal{F}_i^r \in \mathbb{R}^3$ is a continuous feature code, ranging [-1,1], indicating the location of the local texture in the hidden feature space. In our implementation, we use a three-dimensional feature code to represent the local texture information.

For each pixel in the high-resolution grid, we concatenate the feature codes from multiple resolutions to obtain a six-dimensional representation:
\begin{equation}
    \mathcal{F}^r_i = [x, y, z_t, \hat{\mathcal{F}}],
\end{equation}
where $[x, y, z_t]$ represents the pixel coordinate in 3D space, and $\mathcal{F}_i^r$ represents the local texture features from multiple resolutions.

\subsection{Implicit Hashing}
To obtain the implicit feature vectors for a given coordinate in the 6D latent space, we leverage a spatial hash table. The hash table stores trainable feature vectors in a compact manner, allowing us to efficiently retrieve feature codes for neighboring coordinates.

For a given high-resolution coordinate, we find the neighboring locations in the latent space and use the hash table to retrieve their feature vectors:
\begin{equation}
    \{\mathcal{F}_{N_D}^r\} = \text{HashTable}(\mathcal{F}_i),
\end{equation}
where $\{\mathcal{F}_{N_D}^r\}$ represents the feature vectors of the neighboring nodes in the latent space.

We then use a network to predict weights for combining the neighboring features. As we encode the local features using a 6D code (where the first three dimensions indicate the spatial location of the coordinates, and the last three dimensions, predicted by $f^r_\theta$, indicate the coordinates of local texture in the implicit texture codebook, which is unknown), the number of neighbors in the 6D space will be $2^D$ (where $D = 6$ in our implementation). Directly interpolating feature vectors in this high-dimensional space is computationally intensive, and the unknown nature of the feature codebook adds further complexity. To address this, we use a network to predict the weights for each neighbor:
\begin{equation}
    \text{weight}_N^r = f_H(\mathcal{F}_i^r, \mathcal{F}_i^r - \mathcal{F}_i^r(\text{min}), \mathcal{F}_i^r(\text{max}) - \mathcal{F}_i^r),
\end{equation}
where $\mathcal{F}_i^r(\text{min})$ and $\mathcal{F}_i^r(\text{max})$ represent the bottom and top boundaries of the high-dimensional vertex, respectively. The final feature vector at resolution layer $r$ is obtained by a weighted sum of the neighboring feature vectors:
\begin{equation}
    \mathcal{F}_p^r = \sum \text{weight}_N^r \cdot \mathcal{F}_{N_D}^r.
\end{equation}

\subsection{Top-Down Attention for Feature Concatenation}
To generate the final feature vector, we apply a top-down attention mechanism to integrate features from different resolution layers. We start from the top layer and iteratively update the feature vectors using attention weights:
\begin{equation}
    \text{att}_w^{r+1} = f_{att}(\mathcal{F}_p^r),
\end{equation}
\begin{equation}
    \mathcal{F}_p^{r+1} = \text{att}_w^{r+1} \cdot \mathcal{F}_p^{r+1}.
\end{equation}

After applying the attention weights, we concatenate the updated feature vectors from all resolution layers to form the final feature representation:
\begin{equation}
    \mathcal{F}_p = [\mathcal{F}_p^1, \mathcal{F}_p^2, \dots, \mathcal{F}_p^R].
\end{equation}

The final RGB color value for each high-resolution pixel is then predicted using a two-layer MLP:
\begin{equation}
    C_p = f_C(\mathcal{F}_p),
\end{equation}
where $f_C$ is the MLP used to map the concatenated feature vector to the RGB color value.

All networks used in our pipeline are two-layer MLPs with one hidden layer of 64 nodes.

\subsection{Training Details.}
Due to the pixel-based encoding and decoding nature of our method, using a simple mean squared error (MSE) loss often introduces artifacts. These artifacts occur because MSE loss equally penalizes all pixels, causing the network to overlook subtle yet critical details in regions with small reconstruction errors, which can lead to over-smoothing. To address this issue, we introduce a novel Pixel-Error Amplified Loss (PEA-loss).

Specifically, the PEA-loss is designed to amplify reconstruction errors for low-error regions to encourage the model to refine these areas further. The loss is defined as follows:

First, we calculate the per-pixel reconstruction error:
\begin{equation} \label{eqn:lpixel}
    \mathcal{L}_{\text{pixel}} = \text{MSE}(C_p, C_{\text{gt}}),
\end{equation}
where $C_p$ is the predicted RGB value, and $C_{\text{gt}}$ is the ground truth RGB value.
Next, we apply a reconstruction mask to emphasize areas with higher errors. Initially, the reconstruction mask is set to all ones, we then update the reconstruction mask by masking out pixels with low reconstruction error:
\begin{equation}
    M_{\text{recon}} = \begin{cases}
        M_{\text{recon}}, & \text{if } \mathcal{L}_{\text{pixel}} > \tau, \\
        0, & \text{otherwise},
    \end{cases}
\end{equation}
where $\tau$ is a predefined threshold for masking low-error pixels.
The masked loss is then calculated as:
\begin{equation}
    \mathcal{L}_{\text{masked}} = \text{mean}(\mathcal{L}_{\text{pixel}} \cdot M_{\text{recon}}),
\end{equation}

Additionally, we introduce a boosted loss for regions with small errors to further refine details:
\begin{equation}
    \mathcal{L}_{\text{boost}} = f_{\text{boost}}(C_p, C_{\text{gt}}) = \mathcal{L}_{\text{pixel}} + M_{\text{boost}} \cdot \delta,
\end{equation}
with $M_{\text{boost}}$ defined as:
\begin{equation}
    M_{\text{boost}} = \begin{cases}
        1, & \text{if } \mathcal{L}_{\text{pixel}} < \epsilon, \\
        0, & \text{otherwise},
    \end{cases}
\end{equation}
where $\epsilon$ is a small threshold for boosting low-error pixels, and $\delta$ is a small constant added to amplify the loss.

The final PEA-loss is given by:
\begin{equation}
    \mathcal{L}_{\text{PEA}} = \mathcal{L}_{\text{recon}} + \alpha \cdot \mathcal{L}_{\text{boost}}
\end{equation}
where $\alpha$ is the parameter to control the boosting impact.
A detailed analysis of the loss related parameters can be found in our Supplementary material.

\begin{figure*}[t!]
    \centering
    \includegraphics[width=0.85\textwidth]{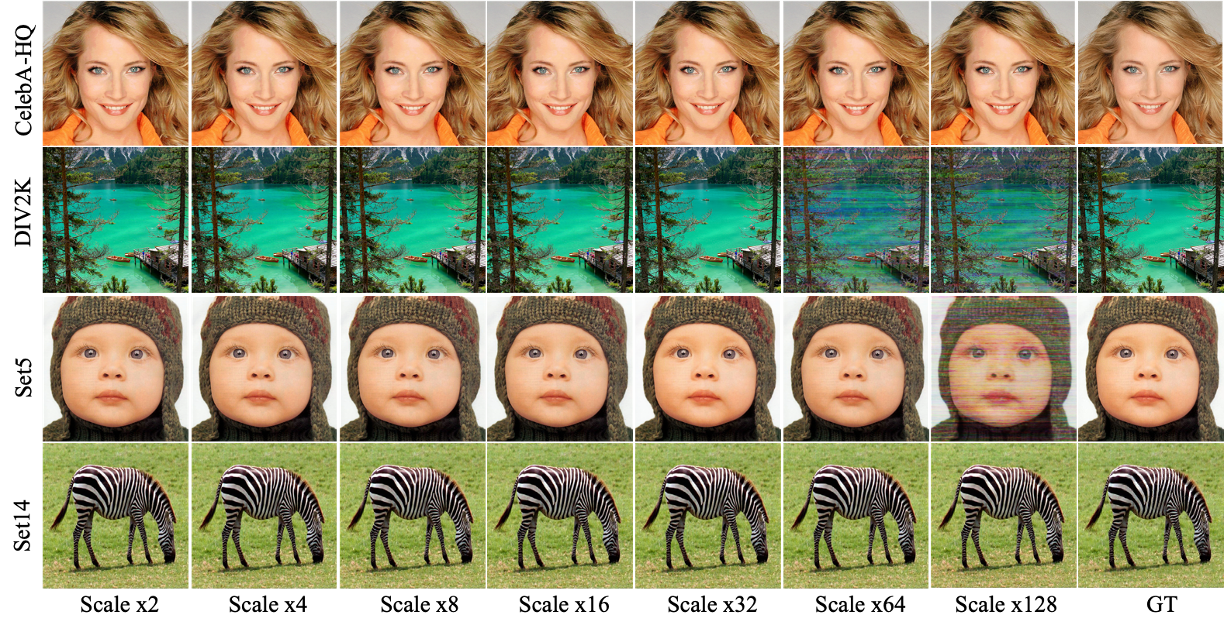}
    \caption{Generalization capability of our method. Trained with an upsampling scale of $\times$32, our method is subsequently applied to super-resolution tasks at scales of $\times$2, $\times$4, $\times$8, $\times$16, $\times$64, and $\times$128. }
    \label{generation}
\end{figure*}

\section{Experiments}

\begin{table*}[t!]
\small
\centering
\begin{tabular}{c|c|cc|cc|cc|cc} 
\toprule
\multirow{2}{*}{Datasets} & \multirow{2}{*}{Methods} & \multicolumn{2}{c|}{Scale x4} & \multicolumn{2}{c|}{Scale x8} & \multicolumn{2}{c|}{Scale x16} & \multicolumn{2}{c}{Scale x32} \\

& & PSNR & SSIM & PSNR & SSIM & PSNR & SSIM & PSNR & SSIM \\
\midrule
\multirow{4}{*}{DIV2K} 
& LIIF & 27.364 & \textbf{0.849} & 24.012 & 0.760 & 21.462 & 0.675 & 19.444 & 0.607 \\
& EDSR & 25.739 & 0.753 & 21.076 & 0.511 & 19.001 & 0.430 & 17.273 & 0.395 \\
& WGSR & 29.188 & 0.820 & 20.187 & 0.517 & 17.957 & 0.438 & 16.074 & 0.400 \\
& Ours & \textbf{27.870} & 0.807 & \textbf{25.930} & \textbf{0.846} & \textbf{26.230} & \textbf{0.859} & \textbf{27.130} & \textbf{0.892} \\
\midrule
\multirow{4}{*}{CelebA-HQ} 
& LIIF & 33.285 & \textbf{0.949} & 30.442 & \textbf{0.928} & 27.433 & 0.913 & 24.256 & 0.887 \\
& EDSR & 31.879 & 0.843 & 27.233 & 0.726 & 24.144 & 0.665 & 21.186 & 0.618 \\
& WGSR & 22.669 & 0.519 & 26.336 & 0.726 & 23.181 & 0.668 & 20.110 & 0.620 \\
& Ours & \textbf{34.205} & 0.923 & \textbf{34.250} & 0.927 & \textbf{34.430} & \textbf{0.932} & \textbf{34.875} & \textbf{0.948} \\
\midrule
\multirow{4}{*}{Set5}
& LIIF & 32.294 & \textbf{0.963} & \textbf{26.982} & \textbf{0.921} & 23.013 & \textbf{0.873} & 20.329 & 0.806 \\
& EDSR & \textbf{33.984} & 0.911 & 25.744 & 0.729 & 22.098 & 0.630 & 19.090 & 0.574 \\
& WGSR & 31.508 & 0.869 & 25.469 & 0.758 & 21.378 & 0.657 & 18.159 & 0.592 \\
& Ours & 22.180 & 0.614 & 22.770 & 0.628 & \textbf{23.390} & 0.671 & \textbf{35.610} & \textbf{0.982} \\
\midrule
\multirow{4}{*}{Set14}
& LIIF & 28.671 & \textbf{0.920} & 24.938 & 0.822 & 22.134 & 0.745 & 19.824 & 0.708 \\
& EDSR & 27.879 & 0.792 & 22.152 & 0.557 & 19.737 & 0.474 & 17.882 & 0.442 \\
& WGSR & 26.689 & 0.716 & 21.568 & 0.585 & 18.749 & 0.499 & 16.821 & 0.462 \\
& Ours & \textbf{30.030} & 0.791 & \textbf{31.780} & \textbf{0.849} & \textbf{32.410} & \textbf{0.851} & \textbf{37.040} & \textbf{0.961} \\
\bottomrule
\end{tabular}
\caption{Quantitative comparison on image datasets including DIV2K, CelebA-HQ, Set5 and Set14. The best result is highlighted in bold.}
\label{imagecompare}
\end{table*}

\begin{figure*}[htbp]
    \centering
    \includegraphics[width=0.7\textwidth]{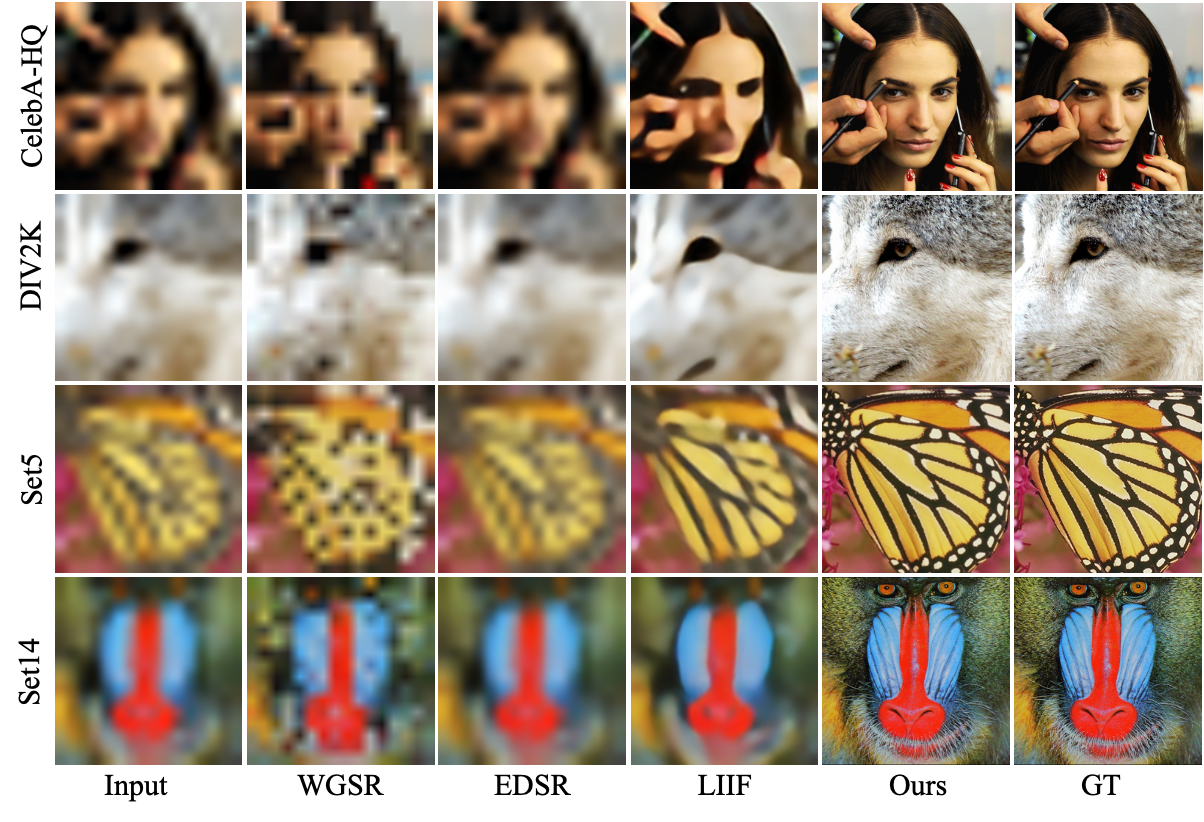}
    \caption{Visual comparison of SOTA in Image SR. From top to bottom, the datasets displayed are CelebA-HQ, DIV2K, Set5, and Set14. The leftmost column shows the input images, while the rightmost column displays the ground truth. The upsampling scale is set to $\times$32.}
    \label{alldata}
\end{figure*}


\begin{table}
  \centering
  \scriptsize 
  \setlength{\tabcolsep}{8pt}  
  \resizebox{\linewidth}{!}{ 
  \begin{tabular}{l|cc|cc}
    \toprule
    \multirow{2}{*}{Method} & \multicolumn{2}{c|}{Vid4} & \multicolumn{2}{c}{GOPRO} \\
    & PSNR & SSIM & PSNR & SSIM\\
    \midrule
    VRT  &  23.780 & 0.672 & 28.800 & 0.854 \\
    VideoINR  &  25.610 & 0.771 & \textbf{29.410} & \textbf{0.867} \\
    Ours  &  \textbf{30.760} & \textbf{0.945} & 27.612 & 0.856\\
    \bottomrule
  \end{tabular}
  }
  \caption{Quantitative comparison on video datasets including Vid4 and GOPRO. Best result in PSNR and SSIM is highlighted in bold.}
  \label{quantitative_on_video}
\end{table}

\begin{table}
\tiny
  \centering
  \scriptsize  
  \setlength{\tabcolsep}{8pt} 
  \resizebox{\linewidth}{!}{ 
  \begin{tabular}{l|cc|cc}
    \toprule
    \multirow{2}{*}{Method} & \multicolumn{2}{c|}{Vid4} & \multicolumn{2}{c}{GOPRO} \\
    & PSNR & SSIM & PSNR & SSIM\\
    \midrule
    NERV  &  \textbf{35.446}&\textbf{0.976} & \textbf{32.028}&\textbf{0.970} \\
    Ours  &  30.760 & 0.945 & 27.612 & 0.856\\
    \bottomrule
  \end{tabular}
  }
  \caption{Quantitative comparison on video datasets including Vid4 and GOPRO. The best result in PSNR and SSIM is highlighted in bold.}
  \label{quantitative_on_nerv}
\end{table}

\subsection{Image Super Resolution}
For our image super-resolution experiments, we evaluated the proposed method on several benchmark datasets, including DIV2K~\cite{DIV2K} from the NTIRE 2017 Challenge~\cite{timofte2017ntire}, CelebA-HQ~\cite{Celebahq}, Set5~\cite{Set5}, and Set14~\cite{Set14}. We trained our model separately on each dataset to assess its performance across different image domains and characteristics.
For DIV2K, we center-cropped each image to $1024 \times 1024$ pixels and  downsampled them to $512 \times 512$ pixels using bicubic interpolation; this $512 \times 512$ resolution served as the HR ground truth for our experiments. For CelebA-HQ, we downsampled the images directly to $512 \times 512$ pixels to obtain the HR images. The Set5 and Set14 datasets were used as provided and adjusted to match our target low-resolution (LR).
We generated the LR inputs with scaling factors of $\times 2$, $\times 4$, $\times 8$, $\times 16$, and $\times 32$. 


To validate the effectiveness of the proposed method, we conducted comprehensive experimental evaluations on several widely used benchmark datasets and compared its performance with state-of-the-art approaches (WGSR \cite{korkmaz2024training}, EDSR \cite{edsr}, LIIF \cite{chen2021liif}). The experimental results demonstrate that our method achieved strong performance in reconstructing high-quality HR images from LR inputs, even at large scaling factors. The results indicated that our model effectively handles high-scale super-resolution, preserving fine details and textures across different datasets.

\begin{figure}[htb]
    \centering
    \includegraphics[width=0.47\textwidth]{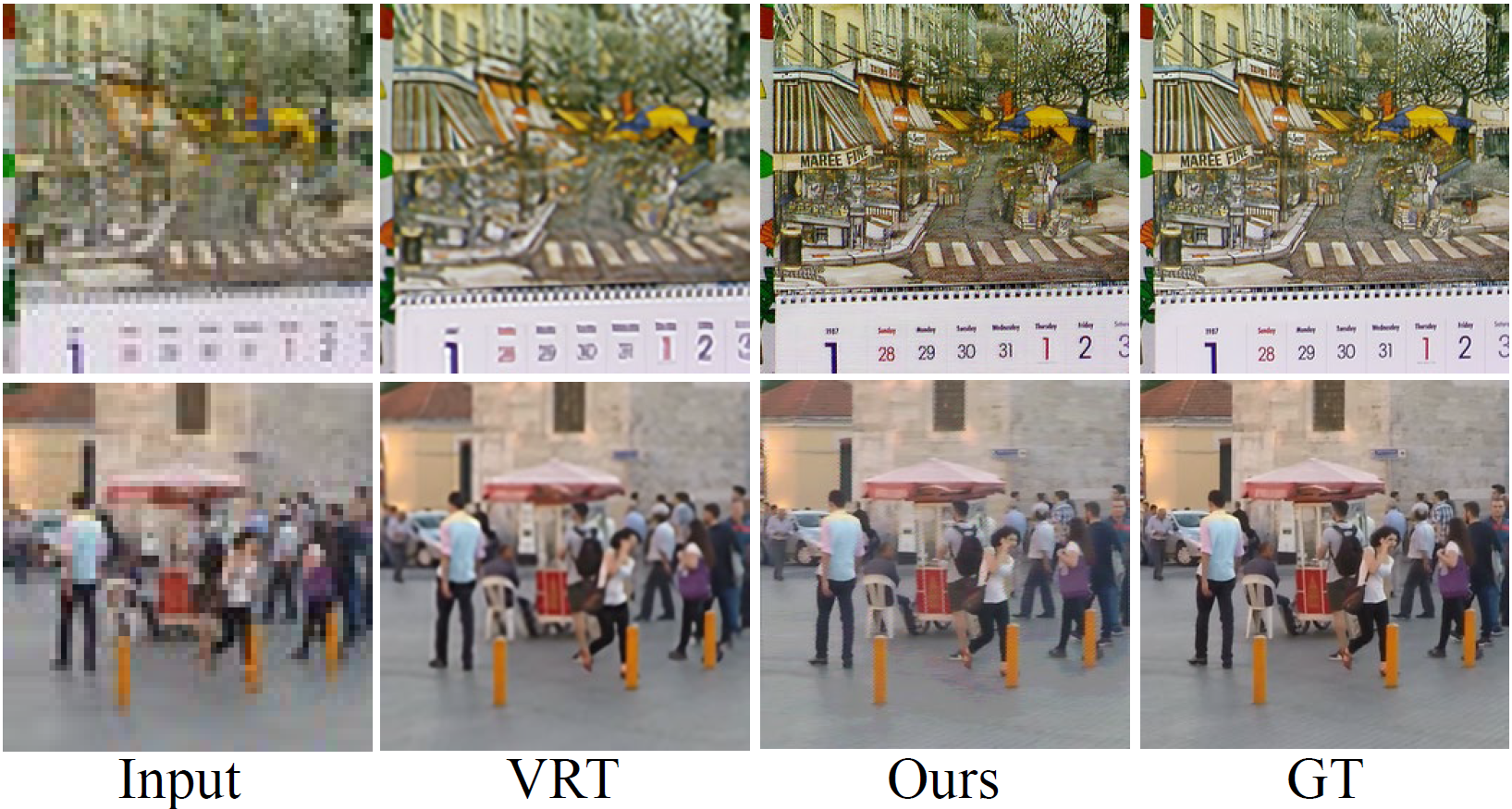}
    \caption{This image shows the results of VRT and Ours. From top to bottom, the datasets displayed are Vid4 and GOPRO datasets.}
    \label{vrt}
\end{figure}

\begin{figure}[htb]
    \centering
    \includegraphics[width=0.47\textwidth]{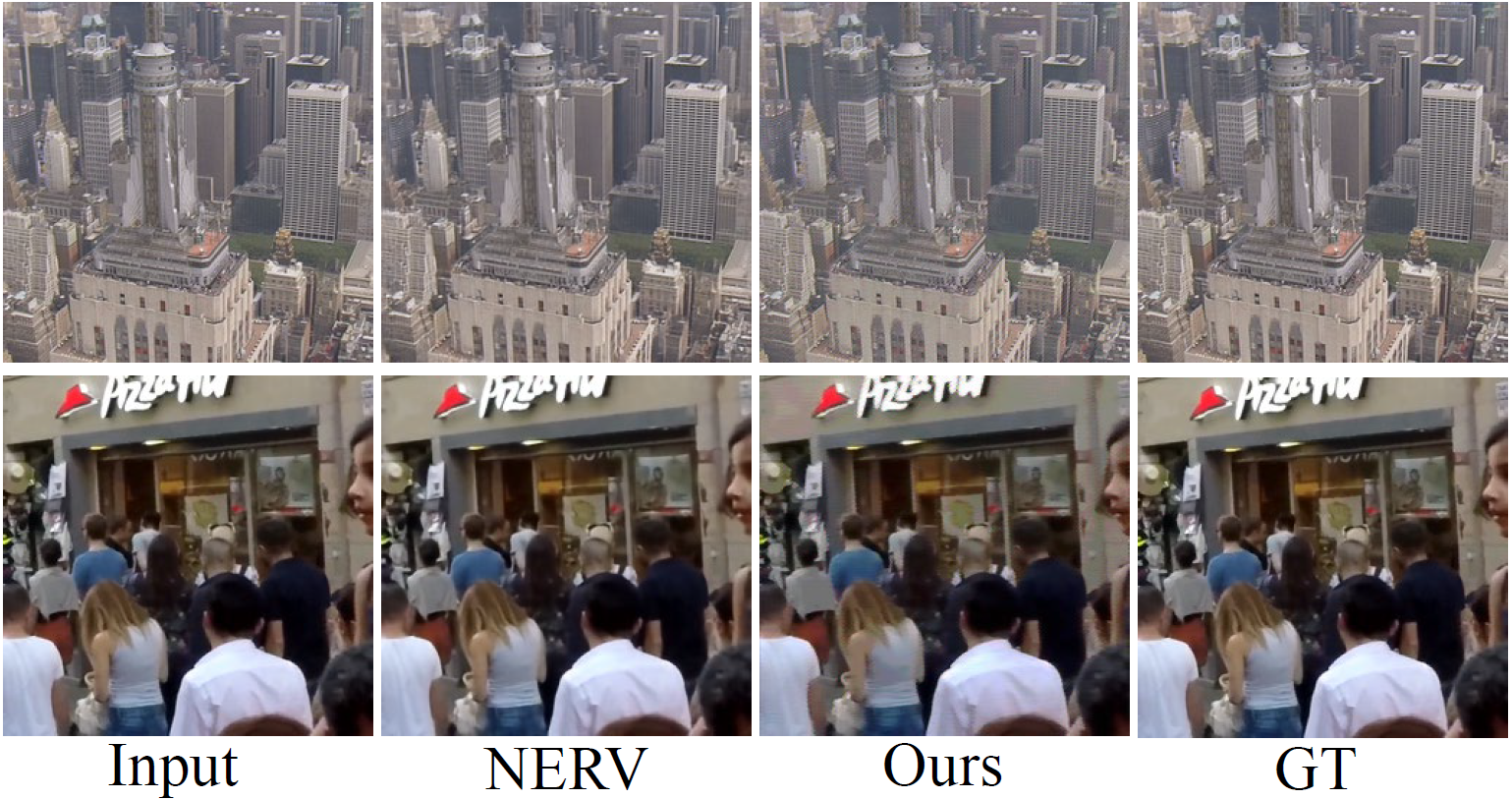}
    \caption{This image shows results of NERV and Ours. From top to bottom, the datasets displayed are Vid4 and GOPRO datasets.}
    \label{nerv}
\end{figure}

\subsection{Quantitative Analysis}
Table\ref{imagecompare} presents the quantitative evaluation results on four benchmark datasets. On the DIV2K dataset, our method achieves a PSNR of 27.130 at ×32 scale, representing a 39.5\% improvement over LIIF, 57.1\% and 68.8\% improvements over EDSR and WGSR, respectively. This performance advantage is consistent across all super-resolution scales, particularly in challenging high-scale scenarios. 
On the CelebA-HQ dataset, at ×32 scale, our method achieves 43.8\% and 6.9\% improvements on PSNR and SSIM respectively, over the best-performing LIIF method. Notably, on the Set14 dataset, the PSNR of our method increased from 30.030 at ×4 to 37.040 at ×32. This counterintuitive result suggests that our approach is better equipped to capture and reconstruct the intrinsic features of images at extremely high super-resolution scale.


\subsection{Qualitative Analysis}
Figure \ref{alldata} showcases a visual comparison of results under ×32 scale. In the face reconstruction tests on the CelebA-HQ dataset, existing methods commonly exhibit significant detail loss. Images generated by WGSR and EDSR display noticeable blurring and structural distortions. While LIIF shows some improvement in preserving overall facial structures, it struggles to capture critical details in areas such as the eyes and facial contours. In contrast, our method not only accurately maintains the overall facial structure but also successfully restores fine features like eyes and hair strands, achieving a visual quality highly consistent with the ground truth.
For handling complex textures, the wolf fur images from DIV2K and butterfly wings from Set5 serve as representative cases. Traditional methods often result in over-smoothing or structural deformation when processing such high-frequency details, leading to a loss of texture information. Our method effectively overcomes these limitations, accurately reconstructing the intricate textures of wolf fur and precisely restoring the complex patterns and structures of butterfly wings. Notably, for the primate facial images in the Set14 dataset, our method demonstrates exceptional color fidelity when handling abrupt transitions between red and blue hues, avoiding common issues of color distortion and edge blurring seen in other approaches.
%
Figure \ref{generation} shows the generalizable ability of our method.We trained our model at a scale of ×32 and subsequently applied it to super-resolution tasks at scales of ×2, ×4, ×8, ×16, ×64, and ×128, demonstrating the excellent generalization performance of our method across all scales overall.

\subsection{Video Super Resolution}
To showcase the effectiveness of our model in video super-resolution, we conducted experiments on the Vid4~\cite{Vid4} and GOPRO~\cite{gopro} video datasets. These datasets contain diverse content and motion patterns, providing a comprehensive evaluation of our method's capability to handle temporal information.
For the experiments, we processed the video frames by resizing the HR video frames to $256 \times 256$ pixels to serve as the ground truth. The low-resolution video sequences were generated by downsampling the HR frames using a scaling factor of $\times 4$ with bicubic interpolation, resulting in LR frames of $64 \times 64$ pixels. Our model was trained to reconstruct the HR frames from the LR inputs, effectively learning to capture both spatial details and temporal consistency across frames. 

We evaluated the performance of our model on video super-resolution tasks by comparing it with state-of-the-art methods, specifically the Video Restoration Transformer (VRT)\cite{liang2022vrt} and VideoINR\cite{chen2022videoinr}. Our experiments were conducted using a scaling factor of $\times 4$, where the low-resolution input frames are $64 \times 64$ pixels, and the high-resolution outputs are $256 \times 256$ pixels. As shown in Figure~\ref{vrt}, our model is capable of generating high-quality results. The quantitative results presented in Table~\ref{quantitative_on_video} indicate that our model achieves the best performance on the VID4 dataset and comparable results on the GOPRO dataset.
Additionally, we compared our method with the video generation model NeRV~\cite{chen2021nerv}, which is trained on high-resolution videos. Despite the differences in training conditions, our model is able to produce high-quality video outputs, as demonstrated in Figure~\ref{nerv}. The comparative results in Table~\ref{quantitative_on_nerv} further confirm the effectiveness of our approach in generating good quality videos.

\subsection{Implementaion Details.} 
All the networks used are two-layer Multi-Layer Perceptrons (MLPs) with ReLU activation functions and hidden layer dimensions of 64 units. The training utilized the Mean Squared Error (MSE) loss function to optimize the model parameters. For evaluation, we considered scaling factors of $\times 2$, $\times 4$, $\times 8$, $\times 16$, and $\times 32$, generating the low-resolution images using bicubic downsampling to simulate different levels of image degradation.
Our model was trained specifically on the $\times 32$ scaling factor for 300 epochs on each dataset, employing a batch size of 32. We used the Adam optimizer with an initial learning rate of $1 \times 10^{-4}$. Learning rate was reduced by a factor of 0.5 every 100 epochs. All experiments were conducted on an NVIDIA A100 GPU.

\subsection{Ablation Studies}
\begin{table}
  \centering
  \tiny 
  \setlength{\tabcolsep}{4pt}
  \resizebox{\linewidth}{!}{ 
  \begin{tabular}{c|c|c|c|c}
    \toprule
    $n\_base\_resolution$ & PSNR & SSIM & Memory & Parameters \\
    \midrule
    n = 2 & \textbf{45.576} & \textbf{0.991} & 3.06G & $4.074 \times 10^8$ \\
    n = 4 & 42.372 & 0.981 & 3.03G & $4.038 \times 10^8$ \\
    n = 8 & 45.080 & 0.988 & 3.02G & $4.030 \times 10^8$ \\
    n = 16 & 44.422 & 0.980 & 3.02G & $4.028 \times 10^8$ \\
    n = 32 & 43.272 & 0.976 & 3.02G & $4.027 \times 10^8$ \\
    \bottomrule
  \end{tabular}
  }
  \caption{Ablation study on the initial grid resolution of the first level. The best result in PSNR and SSIM is highlighted in bold.}
  \label{baseresolution}
\end{table}

\begin{table}
  \centering
  \tiny 
  \setlength{\tabcolsep}{4pt} 
  \resizebox{\linewidth}{!}{ 
  \begin{tabular}{c|c|c|c|c}
    \toprule
    $n\_features\_per\_level$ & PSNR & SSIM & Memory & Parameters \\
    \midrule
    n = 2 & 41.847 & 0.981 & 3.03G & $4.039 \times 10^8$ \\
    n = 4 & \textbf{48.174} & \textbf{0.992} & 6.03G & $8.066 \times 10^8$ \\
    \bottomrule
  \end{tabular}
  }
  \caption{Ablation study on the number of feature codes per level in the hash table. The best result in PSNR and SSIM in bold.}
  \label{featuresperlevel}
\end{table}

We conduct extensive ablation studies to investigate the impact of various architectural choices and hyperparameters on our model's performance on a subset of CelebA-HQ with 100 images and resolution of $128 \times 128$. We evaluate the model using PSNR and SSIM metrics while also considering memory consumption and parameter count. All experiments maintain consistent settings except for the specifically varied component.

\noindent
\textbf{Initial Grid Resolution.}
As shown in Table~\ref{baseresolution}, we experiment with different resolutions ranging from $n=2$ to $n=32$. The results demonstrate that a smaller initial grid resolution ($n=2$) achieves the best performance with PSNR = $45.576$ and SSIM = $0.991$, while maintaining comparable memory usage to other configurations. This suggests that starting with a finer grid resolution allows the model to capture more detailed features from the beginning of the network.

\noindent
\textbf{Feature Codes per Level.}
Table~\ref{featuresperlevel} presents the impact of varying the number of feature codes per level in the hash table. Increasing the feature codes from $n=2$ to $n=4$ leads to a substantial improvement in performance on PSNR for 15.0\% and on SSIM for 1.1\%. However, this enhancement comes at the cost of doubled memory consumption and parameter count. This trade-off suggests that additional feature codes enable better representation capacity but require more computational resources.

\begin{table}
  \centering
  \tiny  
  \setlength{\tabcolsep}{4pt}  
  \resizebox{\linewidth}{!}{ 
  \begin{tabular}{c|c|c|c|c}
    \toprule
    $n\_log2\_hashmap\_size$ & PSNR & SSIM & Memory & Parameters \\
    \midrule
    n = 10 & 13.813 & 0.245 & 0.03G & $1.259 \times 10^6$ \\
    n = 20 & 36.333 & 0.964 & 0.22G & $2.640 \times 10^7$ \\
    n = 22 & 39.858 & 0.977 & 0.78G & $1.019 \times 10^8$ \\
    n = 24 & 42.871 & 0.984 & 3.03G & $4.039 \times 10^8$ \\
    n = 26 & \textbf{44.533} & \textbf{0.988} & 6.03G & $1.612 \times 10^9$ \\
    \bottomrule
  \end{tabular}
  }
  \caption{Ablation study on the size of the hash table per level. The best result in PSNR and SSIM is highlighted in bold.}
  \label{log2hashmapsize}
\end{table}

\noindent
\textbf{Hash Table Size.}
The size of the hash table per level proves to be a crucial factor in model performance, as evidenced in Table~\ref{log2hashmapsize}. We examine different hash table sizes by varying from $10$ to $26$. A clear trend emerges where larger hash tables consistently yield better results, the best performance achieved at $n=26$ with PSNR = $44.533$ and SSIM = $0.988$. However, this improvement comes with a significant increase in memory usage and parameter count, growing from $0.03$G to $6.03$G. Notably, there appears to be a sweet spot around $n=24$, where we achieve good performance (PSNR=$42.871$) with more moderate resource requirements.

\begin{table}
 \centering
 \tiny 
 \setlength{\tabcolsep}{5pt}
 \resizebox{\linewidth}{!}{ 
 \begin{tabular}{c|c|c|c|c}
   \toprule
   $n\_levels$ & PSNR & SSIM & Memory & Parameters \\
   \midrule
   n = 2 & 30.517 & 0.877 & 0.52G & $6.715 \times 10^7$ \\
   n = 4 & 34.835 & 0.924 & 1.02G & $1.347 \times 10^8$ \\
   n = 6 & 36.662 & 0.949 & 1.52G & $2.018 \times 10^8$ \\
   n = 8 & 41.581 & 0.981 & 2.03G & $2.693 \times 10^8$ \\
   n = 10 & \textbf{44.239} & 0.981 & 2.53G & $3.364 \times 10^8$ \\
   n = 12 & 41.117 & 0.976 & 3.03G & $4.039 \times 10^8$ \\
   n = 14 & 42.344 & \textbf{0.988} & 3.53G & $4.710 \times 10^8$ \\
   n = 16 & 43.155 & 0.987 & 4.03G & $5.385 \times 10^8$ \\
   \bottomrule
 \end{tabular}
 }
 \caption{Ablation study on the number of res-layers. The best result in PSNR and SSIM is highlighted in bold.}
 \label{nlevels}
\end{table}

\begin{table}
  \centering
  \tiny  
  \setlength{\tabcolsep}{4pt} 
  \resizebox{\linewidth}{!}{ 
  \begin{tabular}{c|c|c|c|c}
    \toprule
    $n\_finest\_resolution$ & PSNR & SSIM & Memory & Parameters \\
    \midrule
    n = 64 & 31.794 & 0.939 & 6.02G & $4.030 \times 10^8$ \\
    n = 128 & \textbf{39.990} & \textbf{0.976} & 6.04G & $4.039 \times 10^8$ \\
    \bottomrule
  \end{tabular}
  }
  \caption{Ablation study on the number of the final grid resolution of the bottom level. The best result in PSNR and SSIM is highlighted in bold.}
  \label{nfinestres}
  \vspace{-10pt}
\end{table}

\noindent
\textbf{Number of Res-layers.}
Table~\ref{nlevels} investigates the impact of varying the number of res-layers from $2$ to $16$. The results show that performance generally improves with more layers up to a point, with optimal PSNR achieved at $n=10$ layers and best SSIM at $n=14$ layers. Notably, adding more layers beyond $n=10$ doesn't consistently improve performance, suggesting that the model reaches a point of diminishing returns. The memory usage scales linearly with the number of layers, increasing from $0.52$G to $4.03$G.

\noindent
\textbf{Final Grid Resolution.}
The final grid resolution of the bottom level, examined in Table~\ref{nfinestres}, shows that increasing the resolution from $64$ to $128$ significantly improves performance on PSNR with 25.7\% and SSIM with 3.9\%. This improvement comes with only a marginal increase in memory usage, suggesting that a higher final resolution is beneficial for capturing fine details in the output.
\vspace{1em}

These findings provide valuable guidance for configuring the model architecture to achieve the desired balance between performance and computational efficiency.

{
    \small
    \bibliographystyle{ieeenat_fullname}
    \bibliography{main}

\begin{thebibliography}{67}
\providecommand{\natexlab}[1]{#1}
\providecommand{\url}[1]{\texttt{#1}}
\expandafter\ifx\csname urlstyle\endcsname\relax
  \providecommand{\doi}[1]{doi: #1}\else
  \providecommand{\doi}{doi: \begingroup \urlstyle{rm}\Url}\fi

\bibitem[Agustsson and Timofte(2017)]{DIV2K}
Eirikur Agustsson and Radu Timofte.
\newblock {NTIRE} 2017 challenge on single image super-resolution: Dataset and study.
\newblock In \emph{2017 {IEEE} Conference on Computer Vision and Pattern Recognition Workshops, {CVPR} Workshops 2017, Honolulu, HI, USA, July 21-26, 2017}, pages 1122--1131. {IEEE} Computer Society, 2017.

\bibitem[An et~al.(2021)An, Lee, Oh, and Kim]{an2021hyperinr}
Shiqi An, Gwangjin Lee, Tae~Hyun Oh, and Kyoung~Mu Kim.
\newblock Hyperinr: Learning hyper network for continuous image representation with applications to super-resolution.
\newblock In \emph{Advances in Neural Information Processing Systems}, pages 17421--17432, 2021.

\bibitem[Barron et~al.(2021)Barron, Mildenhall, Tancik, Hedman, Martin-Brualla, and Srinivasan]{barron2021mipnerf}
Jonathan~T Barron, Ben Mildenhall, Matthew Tancik, Peter Hedman, Ricardo Martin-Brualla, and Pratul~P Srinivasan.
\newblock Mip-nerf: A multiscale representation for anti-aliasing neural radiance fields.
\newblock In \emph{Proceedings of the IEEE International Conference on Computer Vision}, pages 5855--5864, 2021.

\bibitem[Batzolis et~al.(2022)Batzolis, Pintea, Kraak, and van Gemert]{batzolis2022implicit}
Evangelos Batzolis, Silvia~L. Pintea, Maurice-Jan Kraak, and Jan~C. van Gemert.
\newblock Implicit diffusion models for continuous super-resolution.
\newblock \emph{arXiv preprint arXiv:2207.12527}, 2022.

\bibitem[Bevilacqua et~al.(2012)Bevilacqua, Roumy, Guillemot, and Alberi{-}Morel]{Set5}
Marco Bevilacqua, Aline Roumy, Christine Guillemot, and Marie{-}Line Alberi{-}Morel.
\newblock Low-complexity single-image super-resolution based on nonnegative neighbor embedding.
\newblock In \emph{British Machine Vision Conference, {BMVC} 2012, Surrey, UK, September 3-7, 2012}, pages 1--10. {BMVA} Press, 2012.

\bibitem[Bhat et~al.(2021)Bhat, Kainz, and O'Boyle]{bhat2021deep}
Shahrukh~Athar Bhat, Bernhard Kainz, and Michael O'Boyle.
\newblock Deep implicit functions for image reconstruction.
\newblock In \emph{Proceedings of the IEEE/CVF Conference on Computer Vision and Pattern Recognition}, pages 5649--5658, 2021.

\bibitem[Caballero et~al.(2017)Caballero, Ledig, Aitken, Acosta, Totz, Wang, and Shi]{caballero2017real}
Jose Caballero, Christian Ledig, Andrew Aitken, Alejandro Acosta, Johannes Totz, Zehan Wang, and Wenzhe Shi.
\newblock Real-time video super-resolution with spatio-temporal networks and motion compensation.
\newblock In \emph{Proceedings of the IEEE conference on computer vision and pattern recognition}, pages 4778--4787, 2017.

\bibitem[Chan et~al.(2021)Chan, Wang, Yu, Dong, and Loy]{chan2021basicvsr}
Kelvin~CK Chan, Xintao Wang, Ke Yu, Chao Dong, and Chen~Change Loy.
\newblock Basicvsr: The search for essential components in video super-resolution and beyond.
\newblock In \emph{Proceedings of the IEEE/CVF conference on computer vision and pattern recognition}, pages 4947--4956, 2021.

\bibitem[Chan et~al.(2022)Chan, Zhou, Xu, and Loy]{chan2022basicvsr++}
Kelvin~CK Chan, Shangchen Zhou, Xiangyu Xu, and Chen~Change Loy.
\newblock Basicvsr++: Improving video super-resolution with enhanced propagation and alignment.
\newblock In \emph{Proceedings of the IEEE/CVF conference on computer vision and pattern recognition}, pages 5972--5981, 2022.

\bibitem[Chen et~al.(2022{\natexlab{a}})Chen, Xu, Geiger, Yu, and Su]{chen2022tensorf}
Anpei Chen, Zexiang Xu, Andreas Geiger, Jingyi Yu, and Hao Su.
\newblock Tensorf: Tensorial radiance fields.
\newblock In \emph{Proceedings of the European Conference on Computer Vision}, pages 1--16, 2022{\natexlab{a}}.

\bibitem[Chen et~al.(2022{\natexlab{b}})Chen, Li, Jin, Fan, Yu, and Sun]{chen2022videoinr}
Tao Chen, Jianmin Li, Xuanchong Jin, Haoqiang Fan, Nenghai Yu, and Jian Sun.
\newblock Videoinr: Learning video implicit neural representation for continuous space-time super-resolution.
\newblock In \emph{Proceedings of the IEEE/CVF Conference on Computer Vision and Pattern Recognition}, pages 16105--16114, 2022{\natexlab{b}}.

\bibitem[Chen et~al.(2021{\natexlab{a}})Chen, Liu, and Wang]{chen2021liif}
Yinbo Chen, Sifei Liu, and Xiaolong Wang.
\newblock Learning continuous image representation with local implicit image function.
\newblock In \emph{Proceedings of the IEEE/CVF Conference on Computer Vision and Pattern Recognition}, pages 8628--8638, 2021{\natexlab{a}}.

\bibitem[Chen et~al.(2021{\natexlab{b}})Chen, Wu, and Wang]{chen2021nerv}
Zhengyu Chen, Hexiang Wu, and Yuan-Fang Wang.
\newblock Nerv: Neural representations for videos.
\newblock In \emph{Advances in Neural Information Processing Systems}, 2021{\natexlab{b}}.

\bibitem[Chen et~al.(2024)Chen, Long, Qiu, Yao, Zhou, Luo, and Mei]{chen2024learning}
Zhikai Chen, Fuchen Long, Zhaofan Qiu, Ting Yao, Wengang Zhou, Jiebo Luo, and Tao Mei.
\newblock Learning spatial adaptation and temporal coherence in diffusion models for video super-resolution.
\newblock In \emph{Proceedings of the IEEE/CVF Conference on Computer Vision and Pattern Recognition}, pages 9232--9241, 2024.

\bibitem[Dai et~al.(2019)Dai, Cai, Zhang, Xia, and Zhang]{san}
Tao Dai, Jianrui Cai, Yongbing Zhang, Shu-Tao Xia, and Lei Zhang.
\newblock Second-order attention network for single image super-resolution.
\newblock In \emph{Proceedings of the IEEE/CVF conference on computer vision and pattern recognition}, pages 11065--11074, 2019.

\bibitem[Dong et~al.(2015)Dong, Loy, He, and Tang]{srcnn}
Chao Dong, Chen~Change Loy, Kaiming He, and Xiaoou Tang.
\newblock Image super-resolution using deep convolutional networks.
\newblock \emph{IEEE transactions on pattern analysis and machine intelligence}, 38\penalty0 (2):\penalty0 295--307, 2015.

\bibitem[Dosovitskiy(2020)]{vit}
Alexey Dosovitskiy.
\newblock An image is worth 16x16 words: Transformers for image recognition at scale.
\newblock \emph{arXiv preprint arXiv:2010.11929}, 2020.

\bibitem[Han et~al.(2022)Han, Chen, and Wang]{han2022hnerv}
Rui Han, Zhengyu Chen, and Yuan-Fang Wang.
\newblock Hnerv: A hybrid neural representation for videos.
\newblock In \emph{Proceedings of the IEEE/CVF Conference on Computer Vision and Pattern Recognition}, pages 18966--18975, 2022.

\bibitem[Haris et~al.(2019)Haris, Shakhnarovich, and Ukita]{haris2019recurrent}
Muhammad Haris, Gregory Shakhnarovich, and Norimichi Ukita.
\newblock Recurrent back-projection network for video super-resolution.
\newblock In \emph{Proceedings of the IEEE/CVF conference on computer vision and pattern recognition}, pages 3897--3906, 2019.

\bibitem[He et~al.(2016)He, Zhang, Ren, and Sun]{resnet}
Kaiming He, Xiangyu Zhang, Shaoqing Ren, and Jian Sun.
\newblock Deep residual learning for image recognition.
\newblock In \emph{Proceedings of the IEEE conference on computer vision and pattern recognition}, pages 770--778, 2016.

\bibitem[Huang et~al.(2017)Huang, Wang, and Wang]{huang2017video}
Yan Huang, Wei Wang, and Liang Wang.
\newblock Video super-resolution via bidirectional recurrent convolutional networks.
\newblock \emph{IEEE transactions on pattern analysis and machine intelligence}, 40\penalty0 (4):\penalty0 1015--1028, 2017.

\bibitem[Jo et~al.(2018)Jo, Oh, Kang, and Kim]{jo2018deep}
Younghyun Jo, Seoung~Wug Oh, Jaeyeon Kang, and Seon~Joo Kim.
\newblock Deep video super-resolution network using dynamic upsampling filters without explicit motion compensation.
\newblock In \emph{Proceedings of the IEEE conference on computer vision and pattern recognition}, pages 3224--3232, 2018.

\bibitem[Kappeler et~al.(2016)Kappeler, Yoo, Dai, and Katsaggelos]{kappeler2016video}
Armin Kappeler, Seunghwan Yoo, Qiqin Dai, and Aggelos~K Katsaggelos.
\newblock Video super-resolution with convolutional neural networks.
\newblock \emph{IEEE transactions on computational imaging}, 2\penalty0 (2):\penalty0 109--122, 2016.

\bibitem[Karras et~al.(2018)Karras, Aila, Laine, and Lehtinen]{Celebahq}
Tero Karras, Timo Aila, Samuli Laine, and Jaakko Lehtinen.
\newblock Progressive growing of gans for improved quality, stability, and variation.
\newblock In \emph{6th International Conference on Learning Representations, {ICLR} 2018, Vancouver, BC, Canada, April 30 - May 3, 2018, Conference Track Proceedings}. OpenReview.net, 2018.

\bibitem[Kim et~al.(2016)Kim, Lee, and Lee]{vdsr}
Jiwon Kim, Jung~Kwon Lee, and Kyoung~Mu Lee.
\newblock Accurate image super-resolution using very deep convolutional networks.
\newblock In \emph{Proceedings of the IEEE conference on computer vision and pattern recognition}, pages 1646--1654, 2016.

\bibitem[Korkmaz et~al.(2024)Korkmaz, Tekalp, and Dogan]{korkmaz2024training}
Cansu Korkmaz, A~Murat Tekalp, and Zafer Dogan.
\newblock Training generative image super-resolution models by wavelet-domain losses enables better control of artifacts.
\newblock In \emph{Proceedings of the IEEE/CVF Conference on Computer Vision and Pattern Recognition}, pages 5926--5936, 2024.

\bibitem[Ledig et~al.(2017{\natexlab{a}})Ledig, Theis, Husz{\'a}r, Caballero, Cunningham, Acosta, Aitken, Tejani, Totz, Wang, and Shi]{ledig2017photo}
Christian Ledig, Lucas Theis, Ferenc Husz{\'a}r, Jose Caballero, Andrew Cunningham, Alejandro Acosta, Andrew Aitken, Alykhan Tejani, Johannes Totz, Zehan Wang, and Wenzhe Shi.
\newblock Photo-realistic single image super-resolution using a generative adversarial network.
\newblock In \emph{Proceedings of the IEEE Conference on Computer Vision and Pattern Recognition}, pages 4681--4690, 2017{\natexlab{a}}.

\bibitem[Ledig et~al.(2017{\natexlab{b}})Ledig, Theis, Husz{\'a}r, Caballero, Cunningham, Acosta, Aitken, Tejani, Totz, Wang, et~al.]{srgan}
Christian Ledig, Lucas Theis, Ferenc Husz{\'a}r, Jose Caballero, Andrew Cunningham, Alejandro Acosta, Andrew Aitken, Alykhan Tejani, Johannes Totz, Zehan Wang, et~al.
\newblock Photo-realistic single image super-resolution using a generative adversarial network.
\newblock In \emph{Proceedings of the IEEE conference on computer vision and pattern recognition}, pages 4681--4690, 2017{\natexlab{b}}.

\bibitem[Lee et~al.(2024)Lee, Kim, Cho, and HAN]{lee2024locality}
Doyup Lee, Chiheon Kim, Minsu Cho, and WOOK~SHIN HAN.
\newblock Locality-aware generalizable implicit neural representation.
\newblock \emph{Advances in Neural Information Processing Systems}, 36, 2024.

\bibitem[Lee et~al.(2022)Lee, Kim, Kim, Jang, and Kim]{lee2022inr}
Jaeho Lee, Junyong Kim, Hyeongseok Kim, In-Young Jang, and Junsik Kim.
\newblock Implicit neural representations with learned encoder for image super-resolution.
\newblock \emph{arXiv preprint arXiv:2203.01944}, 2022.

\bibitem[Li et~al.(2020)Li, Tao, Guo, Qi, Lu, and Jia]{li2020mucan}
Wenbo Li, Xin Tao, Taian Guo, Lu Qi, Jiangbo Lu, and Jiaya Jia.
\newblock Mucan: Multi-correspondence aggregation network for video super-resolution.
\newblock In \emph{Computer Vision--ECCV 2020: 16th European Conference, Glasgow, UK, August 23--28, 2020, Proceedings, Part X 16}, pages 335--351. Springer, 2020.

\bibitem[Liang et~al.(2021)Liang, Cao, Sun, Zhang, Van~Gool, and Timofte]{liang2021swinir}
Jingyun Liang, Jiezhang Cao, Guolei Sun, Kai Zhang, Luc Van~Gool, and Radu Timofte.
\newblock Swinir: Image restoration using swin transformer.
\newblock In \emph{Proceedings of the IEEE/CVF international conference on computer vision}, pages 1833--1844, 2021.

\bibitem[Liang et~al.(2022{\natexlab{a}})Liang, Cao, Zhang, Van~Gool, and Timofte]{liang2022vrt}
Jingyun Liang, Jiezhang Cao, Guolei Zhang, Luc Van~Gool, and Radu Timofte.
\newblock Vrt: A video restoration transformer.
\newblock In \emph{Advances in Neural Information Processing Systems}, 2022{\natexlab{a}}.

\bibitem[Liang et~al.(2022{\natexlab{b}})Liang, Fan, Xiang, Ranjan, Ilg, Green, Cao, Zhang, Timofte, and Gool]{liang2022recurrent}
Jingyun Liang, Yuchen Fan, Xiaoyu Xiang, Rakesh Ranjan, Eddy Ilg, Simon Green, Jiezhang Cao, Kai Zhang, Radu Timofte, and Luc~V Gool.
\newblock Recurrent video restoration transformer with guided deformable attention.
\newblock \emph{Advances in Neural Information Processing Systems}, 35:\penalty0 378--393, 2022{\natexlab{b}}.

\bibitem[Liang et~al.(2024)Liang, Cao, Fan, Zhang, Ranjan, Li, Timofte, and Van~Gool]{liang2024vrt}
Jingyun Liang, Jiezhang Cao, Yuchen Fan, Kai Zhang, Rakesh Ranjan, Yawei Li, Radu Timofte, and Luc Van~Gool.
\newblock Vrt: A video restoration transformer.
\newblock \emph{IEEE Transactions on Image Processing}, 2024.

\bibitem[Lim et~al.(2017)Lim, Son, Kim, Nah, and Mu~Lee]{edsr}
Bee Lim, Sanghyun Son, Heewon Kim, Seungjun Nah, and Kyoung Mu~Lee.
\newblock Enhanced deep residual networks for single image super-resolution.
\newblock In \emph{Proceedings of the IEEE conference on computer vision and pattern recognition workshops}, pages 136--144, 2017.

\bibitem[Liu and Sun(2011)]{Vid4}
Ce Liu and Deqing Sun.
\newblock A bayesian approach to adaptive video super resolution.
\newblock In \emph{The 24th {IEEE} Conference on Computer Vision and Pattern Recognition, {CVPR} 2011, Colorado Springs, CO, USA, 20-25 June 2011}, pages 209--216. {IEEE} Computer Society, 2011.

\bibitem[Liu et~al.(2021)Liu, Lin, Cao, Hu, Wei, Zhang, Lin, and Guo]{swint}
Ze Liu, Yutong Lin, Yue Cao, Han Hu, Yixuan Wei, Zheng Zhang, Stephen Lin, and Baining Guo.
\newblock Swin transformer: Hierarchical vision transformer using shifted windows.
\newblock In \emph{Proceedings of the IEEE/CVF international conference on computer vision}, pages 10012--10022, 2021.

\bibitem[Lyu et~al.(2024)Lyu, Wang, Tian, Zou, Dong, Wang, Aviles-Rivero, and Qin]{lyu2024stadnet}
Jun Lyu, Shuo Wang, Yapeng Tian, Jing Zou, Shunjie Dong, Chengyan Wang, Angelica~I Aviles-Rivero, and Jing Qin.
\newblock Stadnet: Spatial-temporal attention-guided dual-path network for cardiac cine mri super-resolution.
\newblock \emph{Medical Image Analysis}, 94:\penalty0 103142, 2024.

\bibitem[Mildenhall et~al.(2020)Mildenhall, Srinivasan, Tancik, Barron, Ramamoorthi, and Ng]{mildenhall2020nerf}
Ben Mildenhall, Pratul~P Srinivasan, Matthew Tancik, Jonathan~T Barron, Ravi Ramamoorthi, and Ren Ng.
\newblock Nerf: Representing scenes as neural radiance fields for view synthesis.
\newblock In \emph{Proceedings of the European Conference on Computer Vision}, pages 405--421, 2020.

\bibitem[M{\"u}ller et~al.(2022)M{\"u}ller, Evans, Schied, and Keller]{mueller2022instant}
Thomas M{\"u}ller, Alex Evans, Christoph Schied, and Alexander Keller.
\newblock Instant neural graphics primitives with a multiresolution hash encoding.
\newblock \emph{ACM Transactions on Graphics}, 41\penalty0 (4):\penalty0 102:1--102:15, 2022.

\bibitem[Nah et~al.(2017)Nah, Kim, and Lee]{gopro}
Seungjun Nah, Tae~Hyun Kim, and Kyoung~Mu Lee.
\newblock Deep multi-scale convolutional neural network for dynamic scene deblurring.
\newblock In \emph{2017 {IEEE} Conference on Computer Vision and Pattern Recognition, {CVPR} 2017, Honolulu, HI, USA, July 21-26, 2017}, pages 257--265. {IEEE} Computer Society, 2017.

\bibitem[Park et~al.(2003)Park, Park, and Kang]{park2003super}
Sung~Cheol Park, Min~Kyu Park, and Moon~Gi Kang.
\newblock Super-resolution image reconstruction: a technical overview.
\newblock \emph{IEEE Signal Processing Magazine}, 20\penalty0 (3):\penalty0 21--36, 2003.

\bibitem[Sajjadi et~al.(2018)Sajjadi, Vemulapalli, and Brown]{sajjadi2018frame}
Mehdi~SM Sajjadi, Raviteja Vemulapalli, and Matthew Brown.
\newblock Frame-recurrent video super-resolution.
\newblock In \emph{Proceedings of the IEEE conference on computer vision and pattern recognition}, pages 6626--6634, 2018.

\bibitem[Shi et~al.(2022)Shi, Gu, Xie, Wang, Yang, and Dong]{shi2022rethinking}
Shuwei Shi, Jinjin Gu, Liangbin Xie, Xintao Wang, Yujiu Yang, and Chao Dong.
\newblock Rethinking alignment in video super-resolution transformers.
\newblock \emph{Advances in Neural Information Processing Systems}, 35:\penalty0 36081--36093, 2022.

\bibitem[Shi et~al.(2015)Shi, Chen, Wang, Yeung, Wong, and Woo]{shi2015convolutional}
Xingjian Shi, Zhourong Chen, Hao Wang, Dit-Yan Yeung, Wai-Kin Wong, and Wang-chun Woo.
\newblock Convolutional lstm network: A machine learning approach for precipitation nowcasting.
\newblock \emph{Advances in neural information processing systems}, 28, 2015.

\bibitem[Sitzmann et~al.(2020)Sitzmann, Martel, Bergman, Lindell, and Wetzstein]{sitzmann2020siren}
Vincent Sitzmann, Julien~N.P. Martel, Alexander~W. Bergman, David~B. Lindell, and Gordon Wetzstein.
\newblock Implicit neural representations with periodic activation functions.
\newblock In \emph{Advances in Neural Information Processing Systems}, pages 7462--7473, 2020.

\bibitem[Soh et~al.(2022)Soh, Park, Cho, Shin, and Han]{soh2022piifgan}
Jaejun Soh, Gwanghyun Park, Sunghyun Cho, Jeonghyeon Shin, and Bohyung Han.
\newblock Piifgan: Periodic implicit image function for general image super-resolution.
\newblock In \emph{Proceedings of the IEEE/CVF Conference on Computer Vision and Pattern Recognition}, pages 18917--18926, 2022.

\bibitem[Sun et~al.(2022)Sun, Chen, Xu, and Geiger]{sun2022direct}
Chao Sun, Yifan Chen, Hao Xu, and Andreas Geiger.
\newblock Directvoxgo: Fast direct voxel grid optimization for radiance fields reconstruction.
\newblock \emph{Proceedings of the IEEE/CVF Conference on Computer Vision and Pattern Recognition}, pages 5459--5469, 2022.

\bibitem[Tancik et~al.(2020)Tancik, Srinivasan, Mildenhall, Fridovich-Keil, Raghavan, Singhal, Ramamoorthi, Barron, and Ng]{tancik2020fourier}
Matthew Tancik, Pratul~P. Srinivasan, Ben Mildenhall, Sara Fridovich-Keil, Neel Raghavan, Utkarsh Singhal, Ravi Ramamoorthi, Jonathan~T. Barron, and Ren Ng.
\newblock Fourier features let networks learn high frequency functions in low dimensional domains.
\newblock In \emph{Advances in Neural Information Processing Systems}, pages 7537--7547, 2020.

\bibitem[Tao et~al.(2017)Tao, Gao, Liao, Wang, and Jia]{tao2017detail}
Xin Tao, Hongyun Gao, Renjie Liao, Jue Wang, and Jiaya Jia.
\newblock Detail-revealing deep video super-resolution.
\newblock In \emph{Proceedings of the IEEE international conference on computer vision}, pages 4472--4480, 2017.

\bibitem[Tian et~al.(2020)Tian, Zhang, Fu, and Xu]{tian2020tdan}
Yapeng Tian, Yulun Zhang, Yun Fu, and Chenliang Xu.
\newblock Tdan: Temporally-deformable alignment network for video super-resolution.
\newblock In \emph{Proceedings of the IEEE/CVF conference on computer vision and pattern recognition}, pages 3360--3369, 2020.

\bibitem[Timofte et~al.(2017)Timofte, Agustsson, Van~Gool, Yang, and Zhang]{timofte2017ntire}
Radu Timofte, Eirikur Agustsson, Luc Van~Gool, Ming-Hsuan Yang, and Lei Zhang.
\newblock Ntire 2017 challenge on single image super-resolution: Methods and results.
\newblock In \emph{Proceedings of the IEEE conference on computer vision and pattern recognition workshops}, pages 114--125, 2017.

\bibitem[Wang et~al.(2019)Wang, Chan, Yu, Dong, and Change~Loy]{wang2019edvr}
Xintao Wang, Kelvin~CK Chan, Ke Yu, Chao Dong, and Chen Change~Loy.
\newblock Edvr: Video restoration with enhanced deformable convolutional networks.
\newblock In \emph{Proceedings of the IEEE/CVF conference on computer vision and pattern recognition workshops}, pages 0--0, 2019.

\bibitem[Wu et~al.(2024)Wu, Yang, Sun, Zhang, Li, and Zhang]{wu2024seesr}
Rongyuan Wu, Tao Yang, Lingchen Sun, Zhengqiang Zhang, Shuai Li, and Lei Zhang.
\newblock Seesr: Towards semantics-aware real-world image super-resolution.
\newblock In \emph{Proceedings of the IEEE/CVF conference on computer vision and pattern recognition}, pages 25456--25467, 2024.

\bibitem[Xu et~al.(2021)Xu, Xu, Li, Wang, Sun, and Cheng]{xu2021temporal}
Gang Xu, Jun Xu, Zhen Li, Liang Wang, Xing Sun, and Ming-Ming Cheng.
\newblock Temporal modulation network for controllable space-time video super-resolution.
\newblock In \emph{Proceedings of the IEEE/CVF conference on computer vision and pattern recognition}, pages 6388--6397, 2021.

\bibitem[Xu et~al.(2023)Xu, Hu, Zhu, Dou, Dai, Qiao, and Heng]{xu2023video}
Jiaqi Xu, Xiaowei Hu, Lei Zhu, Qi Dou, Jifeng Dai, Yu Qiao, and Pheng-Ann Heng.
\newblock Video dehazing via a multi-range temporal alignment network with physical prior.
\newblock In \emph{Proceedings of the IEEE/CVF Conference on Computer Vision and Pattern Recognition}, pages 18053--18062, 2023.

\bibitem[Xue et~al.(2019)Xue, Chen, Wu, Wei, and Freeman]{xue2019video}
Tianfan Xue, Baian Chen, Jiajun Wu, Donglai Wei, and William~T Freeman.
\newblock Video enhancement with task-oriented flow.
\newblock \emph{International Journal of Computer Vision}, 127:\penalty0 1106--1125, 2019.

\bibitem[Yang et~al.(2025)Yang, He, Ma, and Zhang]{yang2025motion}
Xi Yang, Chenhang He, Jianqi Ma, and Lei Zhang.
\newblock Motion-guided latent diffusion for temporally consistent real-world video super-resolution.
\newblock In \emph{European Conference on Computer Vision}, pages 224--242. Springer, 2025.

\bibitem[Yi et~al.(2019)Yi, Wang, Jiang, Jiang, and Ma]{yi2019progressive}
Peng Yi, Zhongyuan Wang, Kui Jiang, Junjun Jiang, and Jiayi Ma.
\newblock Progressive fusion video super-resolution network via exploiting non-local spatio-temporal correlations.
\newblock In \emph{Proceedings of the IEEE/CVF international conference on computer vision}, pages 3106--3115, 2019.

\bibitem[Yi et~al.(2021)Yi, Wang, Jiang, Jiang, Lu, Tian, and Ma]{yi2021omniscient}
Peng Yi, Zhongyuan Wang, Kui Jiang, Junjun Jiang, Tao Lu, Xin Tian, and Jiayi Ma.
\newblock Omniscient video super-resolution.
\newblock In \emph{Proceedings of the IEEE/CVF international conference on computer vision}, pages 4429--4438, 2021.

\bibitem[Yu et~al.(2022)Yu, Fridovich-Keil, Tancik, Ling, Liao, Trevithick, Boyd, Prabhakaran, Ramamoorthi, Kanazawa, and Barron]{yu2022plenoxels}
Alex Yu, Sara Fridovich-Keil, Matthew Tancik, Jonathan Ling, Qinhong Liao, Andrew Trevithick, Steven Boyd, Vidya Prabhakaran, Ravi Ramamoorthi, Angjoo Kanazawa, and Jonathan~T Barron.
\newblock Plenoxels: Radiance fields without neural networks.
\newblock \emph{Proceedings of the IEEE/CVF Conference on Computer Vision and Pattern Recognition}, pages 5501--5510, 2022.

\bibitem[Yu et~al.(2024)Yu, Gu, Li, Hu, Kong, Wang, He, Qiao, and Dong]{yu2024scaling}
Fanghua Yu, Jinjin Gu, Zheyuan Li, Jinfan Hu, Xiangtao Kong, Xintao Wang, Jingwen He, Yu Qiao, and Chao Dong.
\newblock Scaling up to excellence: Practicing model scaling for photo-realistic image restoration in the wild.
\newblock In \emph{Proceedings of the IEEE/CVF Conference on Computer Vision and Pattern Recognition}, pages 25669--25680, 2024.

\bibitem[Zeyde et~al.(2010)Zeyde, Elad, and Protter]{Set14}
Roman Zeyde, Michael Elad, and Matan Protter.
\newblock On single image scale-up using sparse-representations.
\newblock In \emph{Curves and Surfaces - 7th International Conference, Avignon, France, June 24-30, 2010, Revised Selected Papers}, pages 711--730. Springer, 2010.

\bibitem[Zhang et~al.(2018{\natexlab{a}})Zhang, Li, Li, Wang, Zhong, and Fu]{rcan}
Yulun Zhang, Kunpeng Li, Kai Li, Lichen Wang, Bineng Zhong, and Yun Fu.
\newblock Image super-resolution using very deep residual channel attention networks.
\newblock In \emph{Proceedings of the European conference on computer vision (ECCV)}, pages 286--301, 2018{\natexlab{a}}.

\bibitem[Zhang et~al.(2018{\natexlab{b}})Zhang, Li, Li, Wang, Zhong, and Fu]{zhang2018rcan}
Yulun Zhang, Kunpeng Li, Kai Li, Lichen Wang, Bineng Zhong, and Yun Fu.
\newblock Image super-resolution using very deep residual channel attention networks.
\newblock In \emph{Proceedings of the European Conference on Computer Vision}, pages 286--301, 2018{\natexlab{b}}.

\bibitem[Zhou et~al.(2024)Zhou, Yang, Wang, Luo, and Loy]{zhou2024upscale}
Shangchen Zhou, Peiqing Yang, Jianyi Wang, Yihang Luo, and Chen~Change Loy.
\newblock Upscale-a-video: Temporal-consistent diffusion model for real-world video super-resolution.
\newblock In \emph{Proceedings of the IEEE/CVF Conference on Computer Vision and Pattern Recognition}, pages 2535--2545, 2024.

\end{thebibliography}
}


\end{document}


\maketitle


\section*{Abstract}

This supplementary material provides extended details and experiments to complement the main text. We present:
\noindent 1) A comprehensive ablation study on our network architecture and training strategy. 
\noindent 2) Additional visual comparison results, including a video comparison available in the supplementary package. 
\noindent 3) A comparison of the reconstruction capabilities of our method with NeRV \cite{chen2021nerv}. 
These additions aim to provide deeper insights into the effectiveness and robustness of our proposed approach.
\section{Ablation Study on Network Design and Training Strategy}
We conducted additional ablation studies to evaluate the impact of various network architectural choices and training strategies on our model's performance. These experiments were performed using a subset of 100 images from the CelebA-HQ dataset, focusing on a $\times 4$ super-resolution task with low-resolution inputs of $32 \times 32$ pixels and target high-resolution outputs of $128 \times 128$ pixels, which is the same validation set as we used in the main paper.

\subsection{Network Design}

\paragraph{MLP Structure:} The MLPs used in our network incorporate ReLU as the activation function for all hidden layers. The choice of activation function for the output layer depends on the intended purpose of the MLP. Specifically, if the MLP is used to predict weights, for instance, $f_H$ and $f_{att}$  
we utilize the sigmoid activation function to ensure output values remain between 0 and 1. For MLPs designed to predict feature codes, such as $f^r_\theta$ 
a tanh activation function is used to represent the feature information effectively.

\paragraph{Top-Down Attention Mechanism:} To assess the contribution of the top-down attention mechanism, we performed an ablation study by removing the attention component from the feature concatenation process. Without the attention mechanism, features from different resolution layers were directly concatenated without any prioritization. The results, shown in Table \ref{attention}, indicate a significant drop in both PSNR and SSIM. This is likely due to the model's reduced ability to effectively integrate information from multiple resolutions, leading to less refined feature representations and poorer texture consistency.

\begin{table}[h!]
  \centering
  \tiny  
  \setlength{\tabcolsep}{4pt} 
  \resizebox{0.6\linewidth}{!}{ 
  \begin{tabular}{ccc}
    \toprule
     & PSNR & SSIM \\
    \midrule
    No Attention & 36.77 & 0.86\\
    Attention & \textbf{42.57} & \textbf{0.94}\\
    \bottomrule
  \end{tabular}
  }
  \caption{Ablation study on the effect of the top-down attention mechanism.}
  \label{attention}
\end{table}

\subsection{Training Strategy}

\paragraph{Pixel-Error Amplified Loss (PEA-Loss):} We evaluated the effectiveness of our proposed Pixel-Error Amplified Loss ($\mathcal{L}_{\text{PEA}}$) by conducting an ablation study in which the model was trained using only the standard per-pixel Mean Squared Error (MSE) loss, denoted as $\mathcal{L}_{\text{pixel}}$. Table \ref{loss_fn} shows that the model trained with $\mathcal{L}_{\text{PEA}}$ achieved significantly better results in terms of PSNR and SSIM compared to the MSE-only version. The PEA-loss amplifies subtle reconstruction errors, allowing the model to focus on refining regions that would otherwise be neglected by the standard MSE loss, ultimately boosting performance.

\begin{table}[h!]
  \centering
  \tiny  
  \setlength{\tabcolsep}{8pt} 
  \resizebox{0.7\linewidth}{!}{ 
  \begin{tabular}{ccc}
    \toprule
     & PSNR & SSIM \\
    \midrule
    $\mathcal{L}_{\text{pixel}}$ & 39.89 & 0.89\\
    $\mathcal{L}_{\text{PEA}}$ & \textbf{42.57} & \textbf{0.94}\\
    \bottomrule
  \end{tabular}
  }
  \caption{Ablation study on the effect of the $\mathcal{L}_{\text{PEA}}$ loss function.}
  \label{loss_fn}
\end{table}

\paragraph{Maximum Number of Pixels per Batch ($N_{\text{max}}$):} 
We conducted an ablation study to evaluate the impact of the maximum number of pixels processed per batch during training. This parameter, denoted as $N_{\text{max}}$, determines the number of pixels updated at each step and directly affects the network's ability to learn implicit structures:
\begin{equation}
    N_{\text{pixels}} = (H \times W)/{N_{\text{max}}},
\end{equation}
where H and W refer to the height and width of the image, respectively.
Table \ref{pixelnum} presents the results of varying $N_{\text{max}}$. Increasing $N_{\text{max}}$ reduces the overall quality of the reconstructed images, as updating based on a larger subset of pixels can lead to incomplete feature learning, thereby breaking the implicit representation of the image content.

\begin{table}[h!]
  \centering
  \small  
  \setlength{\tabcolsep}{4pt} 
  \begin{tabular}{ccccccc}
    \toprule
    $N_{\text{max}}$ & 1 & 2 & 4 & 8 & 16 & 32 \\
    \midrule
    PSNR & \textbf{42.57} & 34.48 & 36.59 & 26.68  & 10.87  & 10.87 \\
    SSIM & \textbf{0.94} & 0.86 & 0.96 & 0.84 & 0.28  & 0.28  \\
    \bottomrule
  \end{tabular}
  \caption{Ablation study on the effect of the number of pixels updated per batch during training. The best results in PSNR and SSIM are highlighted in bold}
  \label{pixelnum}
\end{table}

\section{Visual Comparison}
\subsection{Image Super-Resolution}
\begin{figure*}[t!]
    \centering
    \includegraphics[width=1\textwidth]{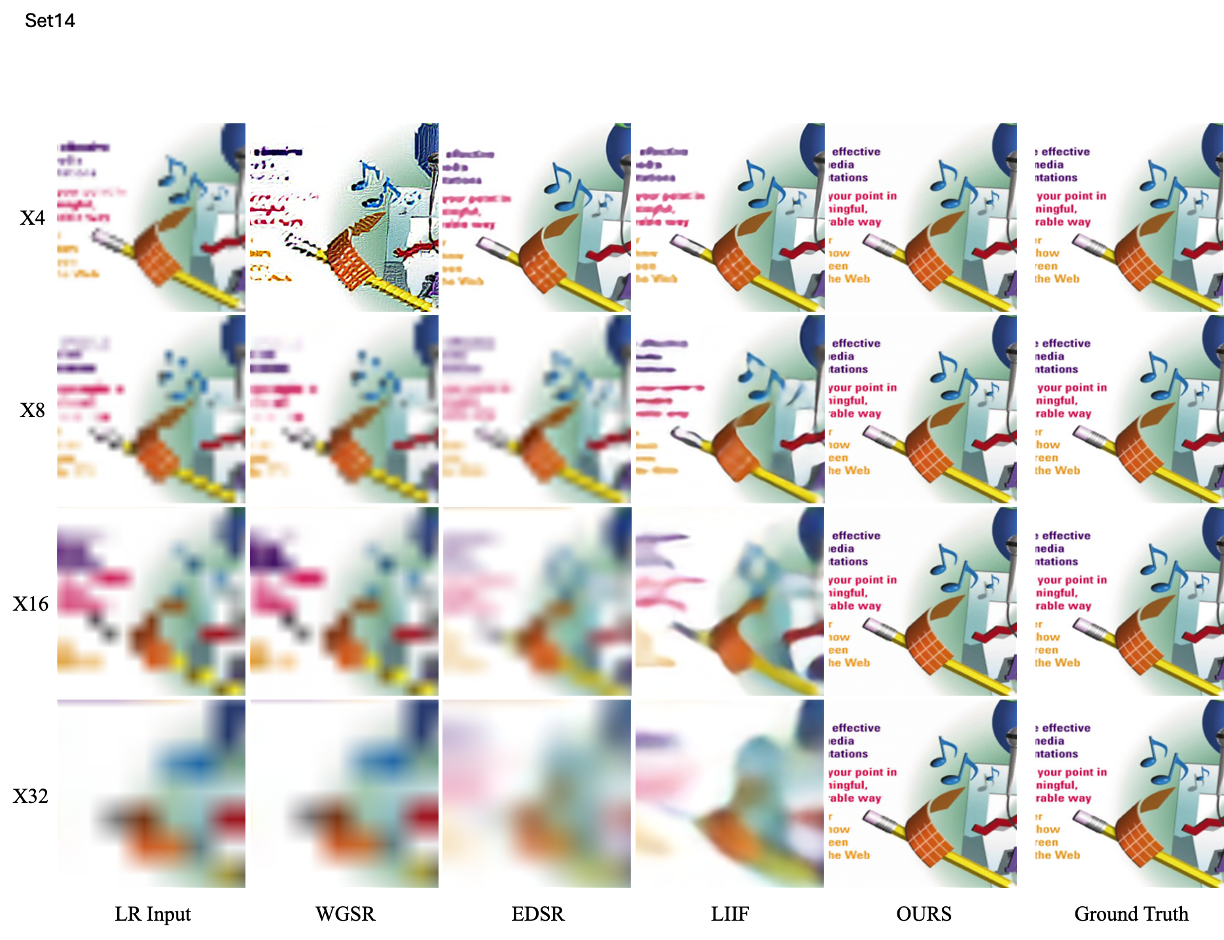}
    \caption{Visual Comparison results on SET14 at different resolution scales.}
    \label{set14}
\end{figure*}

Here, we present additional visual comparisons of our model's performance against other state-of-the-art (SOTA) methods, including WGSR~\cite{korkmaz2024training}, EDSR~\cite{edsr}, and LIIF~\cite{chen2021liif}. The evaluation is conducted on Set14 \cite{Set14}, DIV2K \cite{timofte2017ntire}, and CelebA-HQ \cite{Celebahq} datasets, which encompass a variety of image textures and content types. 
Our comparisons focus on image super-resolution tasks at scaling factors of $\times 4$, $\times 8$, $\times 16$, and $\times 32$, covering moderate and extreme upscaling scenarios. The results demonstrate the capability of our model to handle significant resolution enhancements while preserving fine details and producing visually appealing outputs. Results in Figure \ref{set14} show that other models could not recover the texts in the images as the resolution scales grow, resulting in oversmooth results, especially at high-resolution scales. 

\begin{figure*}[t!]
    \centering
    \includegraphics[width=1\textwidth]{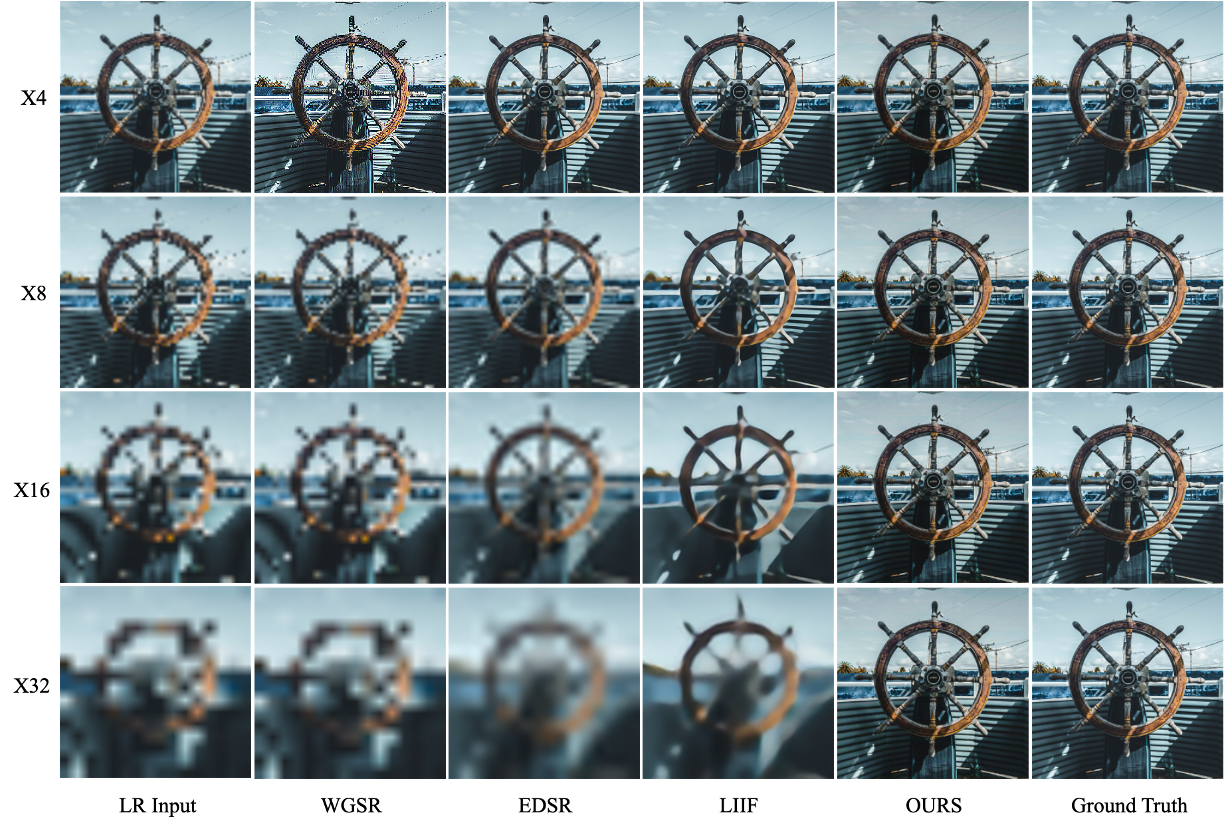}
    \caption{Visual Comparison results on DIV2K dataset at different resolution scales.}
    \label{div2k}
\end{figure*}

\begin{figure*}[t!]
    \centering
    \includegraphics[width=1\textwidth]{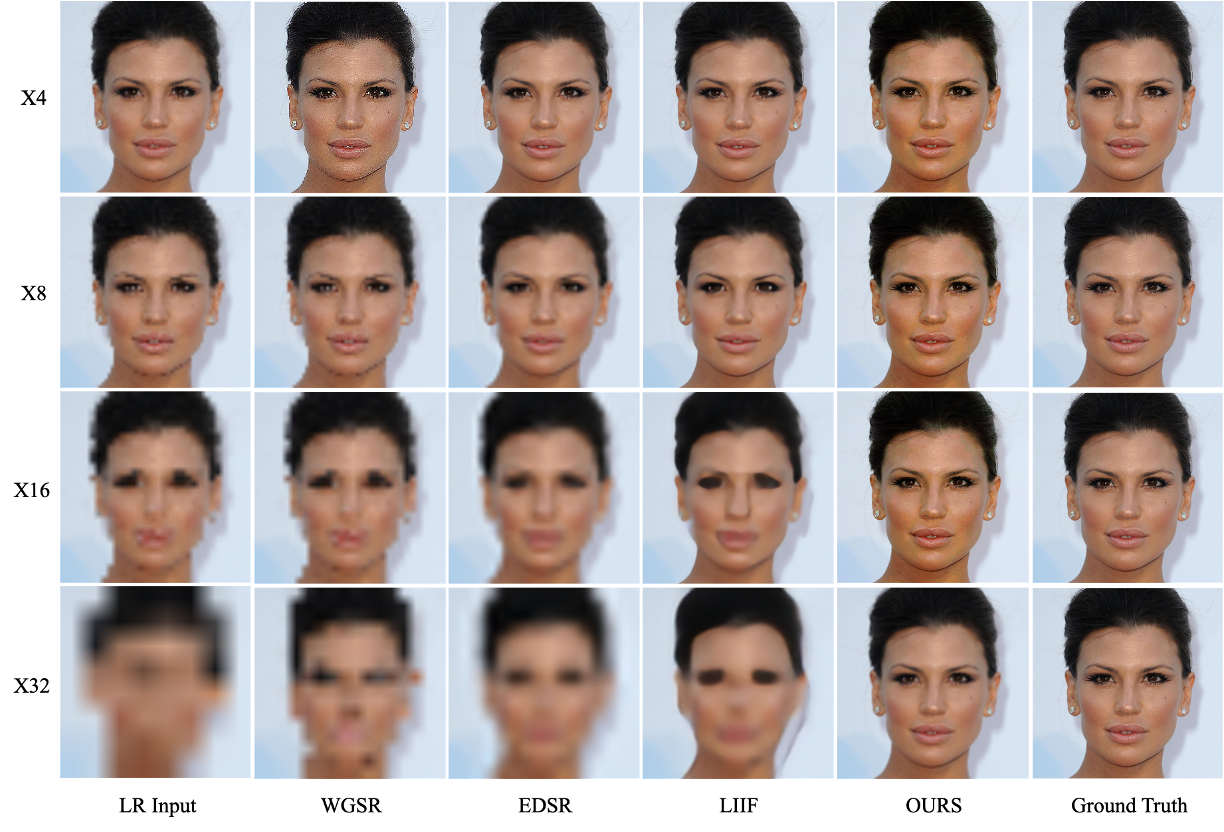}
    \caption{Visual Comparison results on CELEBA-HQ dataset at different resolution scales.}
    \label{celeba}
\end{figure*}

The comparison on DIV2K in Figure \ref{div2k} also demonstrates that our model consistently outperforms other methods in preserving fine details and textures, closely matching the ground truth even at extreme scale factors like $\times 16$ and $\times 32$. Also, as shown in Figure \ref{celeba}, other  models exhibit significant blurring at large scales. This highlights the robustness of our approach in handling challenging image super-resolution tasks.


\begin{figure*}
    \centering
    \includegraphics[width=.9\textwidth]{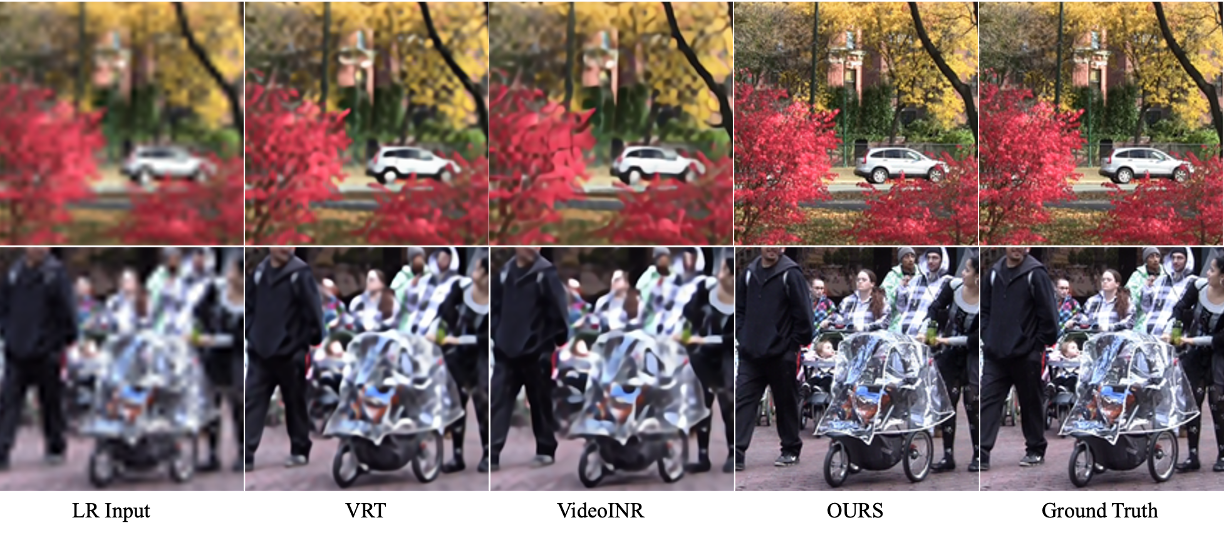}
    \caption{Visual Comparison results on VID4 video dataset at different resolution scales.}
    \label{vid4_sr}
\end{figure*}

\begin{figure*}
    \centering
    \includegraphics[width=1\textwidth]{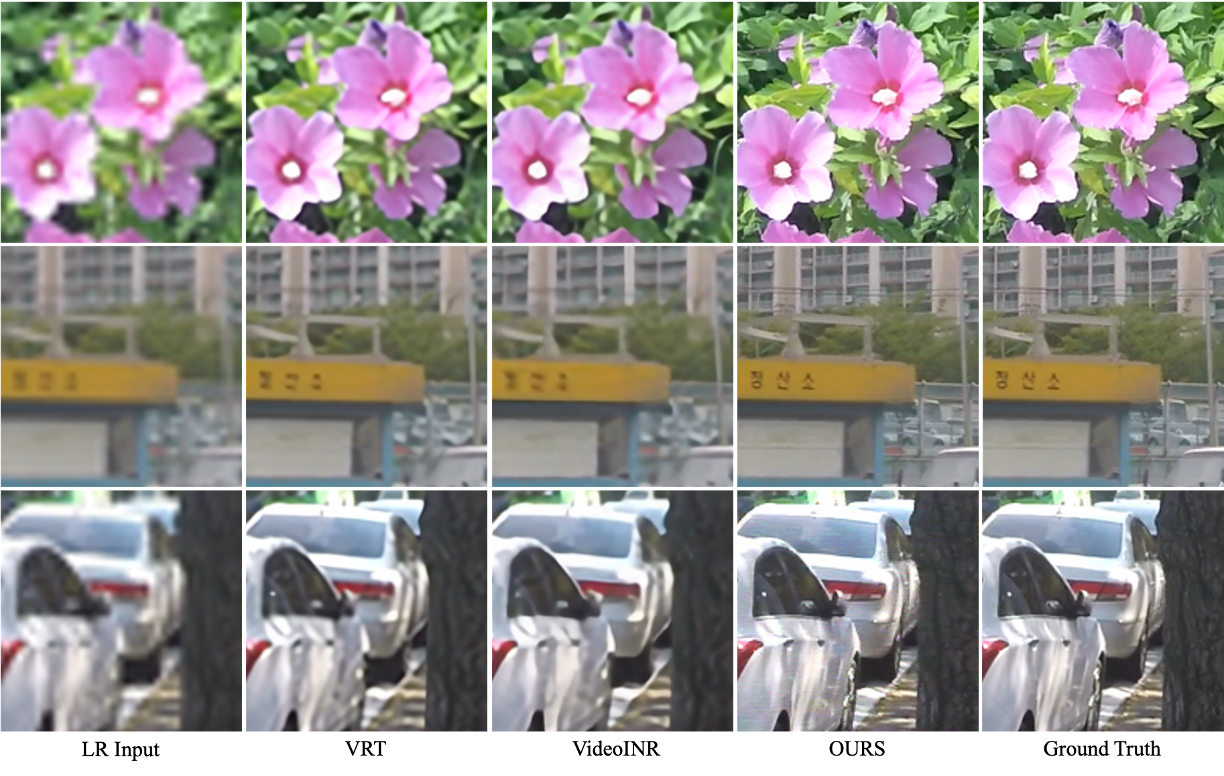}
    \caption{Visual Comparison results on GOPRO video dataset at different resolution scales.}
    \label{gopro_sr}
\end{figure*}

\subsection{Video Super-Resolution}
We show more comparison results on video super-resolution tasks. We also show how robust our model is beyond image super-resolution to video super-resolution. We compared with SOTA methods,  VRT \cite{liang2022vrt} and VideoINR \cite{chen2022videoinr} on Vid4~\cite{Vid4} and GOPRO~\cite{gopro} video datasets. Our comparisons focus on video super-resolution tasks at  $\times 4$ scaling factor with a high resolution of 256x256.

For the Vid4 dataset, our model outperforms other methods by effectively preserving fine textures and intricate details. As shown in Figure \ref{vid4_sr}, the texture of the leaves is significantly more defined in our model and closely resembles the ground truth compared to VRT and VideoINR, where the details are either overly smoothed or partially lost.
For the GOPRO dataset, OURS maintains sharper visual fidelity in challenging regions, as seen in Figure \ref{gopro_sr}. The flower petals retain their vibrant texture and color while competing methods struggle with some blurs. In the second row, the yellow sign with text appears legible and closely matches the ground truth in ours, whereas other models lose the sharpness and clarity of the text. These results emphasize the robustness of our approach in preserving fine details and reconstructing textures across diverse video datasets and scaling factors.

\section{Reconstruction Ability Compared with NERV}
\begin{figure*}
    \centering
    \includegraphics[width=.85\textwidth]{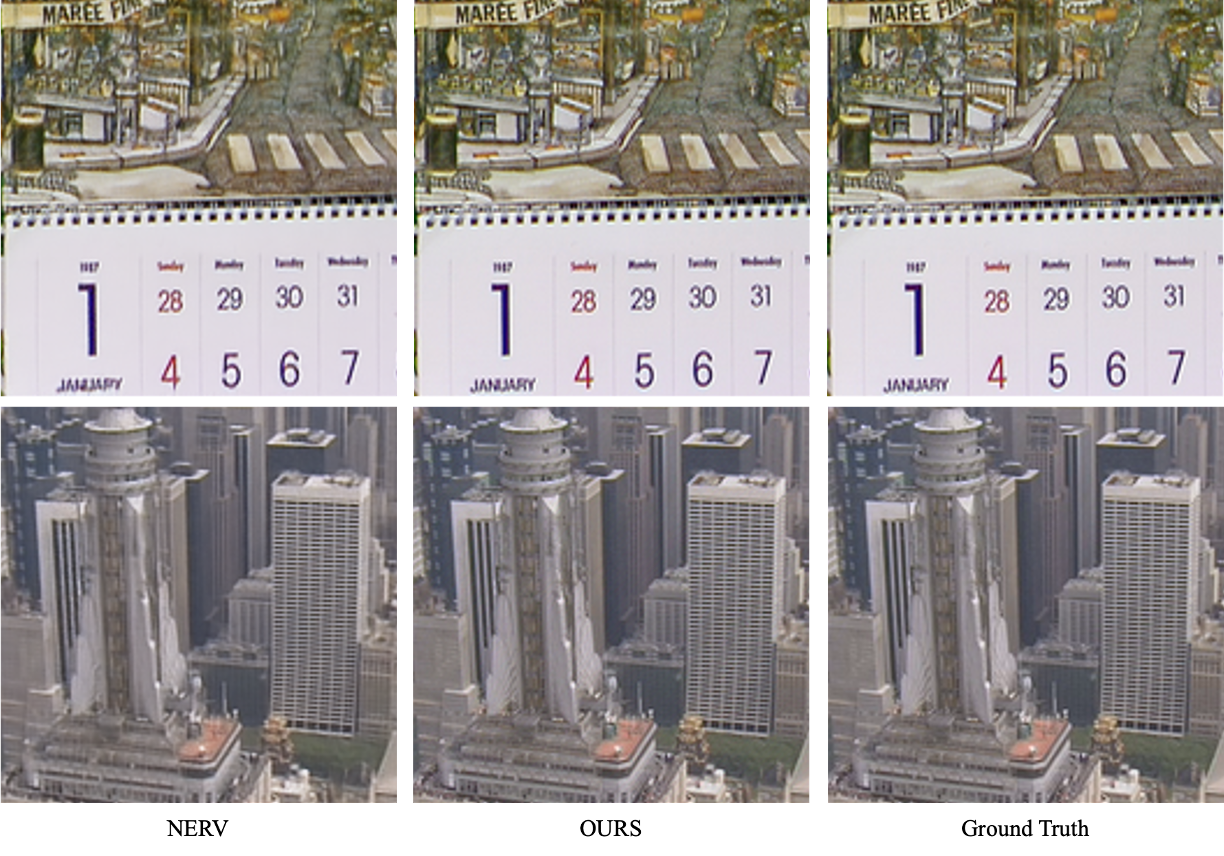}
    \caption{Visual comparison of video reconstruction on VID4 video dataset.}
    \label{div4_recon}
\end{figure*}

\begin{figure*}
    \centering
    \includegraphics[width=.85\textwidth]{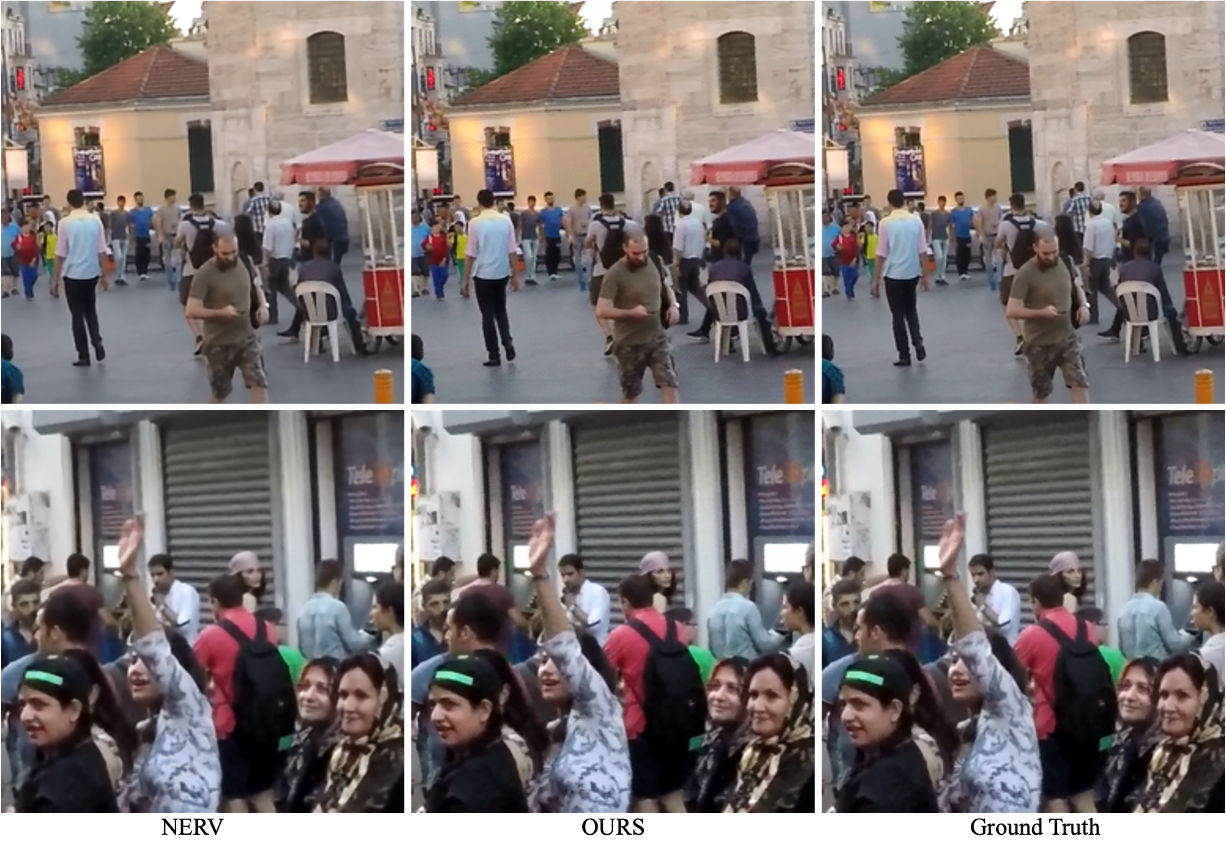}
    \caption{Visual comparison of video reconstruction on GOPRO video dataset.}
    \label{gopro_recon}
\end{figure*}

In addition to image and video super-resolution, our model demonstrates exceptional capabilities in video reconstruction tasks. To evaluate its reconstruction performance, we compared our model with NeRV, a state-of-the-art approach designed specifically for neural video representations, using the GOPRO and VID4 video datasets. PSNR and SSIM were used as evaluation metrics to assess reconstruction quality. As shown in Table \ref{recon_on_nerv}, our model achieves significantly higher PSNR and SSIM values on the VID4 dataset compared to NeRV. Furthermore, Figures \ref{div4_recon} and \ref{gopro_recon} illustrate that our model generates visually sharper and more detailed reconstructions, closely aligning with the ground truth. These results demonstrate that our model not only preserves intricate details effectively but also achieves comparable or superior reconstruction quality to NeRV, showcasing its versatility in addressing a broader range of video-related tasks beyond super-resolution.

\begin{table}
\tiny
  \centering
  \scriptsize  
  \setlength{\tabcolsep}{8pt} 
  \resizebox{\linewidth}{!}{ 
  \begin{tabular}{l|cc|cc}
    \toprule
    \multirow{2}{*}{Method} & \multicolumn{2}{c|}{Vid4} & \multicolumn{2}{c}{GOPRO} \\
    & PSNR & SSIM & PSNR & SSIM\\
    \midrule
    NERV  &  {32.03}&{0.97} & \textbf{35.68}&\textbf{0.98} \\
    Ours  &  \textbf{46.80} & \textbf{0.99} & {34.30} & {0.92}\\
    \bottomrule
  \end{tabular}
  }
  \caption{Quantitative comparison on video datasets including Vid4 and GOPRO. The best results in PSNR and SSIM are highlighted in bold.}
  \label{recon_on_nerv}
\end{table}

{
    \small
    \bibliographystyle{ieeenat_fullname}
    \bibliography{main}
}
